\documentclass[10pt,twocolumn,letterpaper]{article}

\usepackage{iccv}
\usepackage{times}
\usepackage{epsfig}
\usepackage{graphicx}
\usepackage{amsmath}
\usepackage{amssymb}
\usepackage{enumitem}
\usepackage{capt-of}
\usepackage{multirow}
\usepackage[normalem]{ulem}
\usepackage[dvipsnames]{xcolor}

\usepackage{algorithm}
\usepackage[noend]{algorithmic}

\usepackage[pagebackref=true,breaklinks=true,letterpaper=true,colorlinks,bookmarks=false]{hyperref}

\iccvfinalcopy %

\useunder{\uline}{\ul}{}

\begin{document}

\title{Learning Generative Models of Textured 3D Meshes from Real-World Images}

\author{Dario Pavllo \qquad Jonas Kohler \qquad Thomas Hofmann \qquad Aurelien Lucchi\vspace{2mm}\\
Department of Computer Science\\
ETH Zurich
}

\renewcommand{\paragraph}[1]{\noindent {\bf #1.}}
\renewcommand*{\figureautorefname}{Fig.}
\renewcommand*{\sectionautorefname}{sec.}
\renewcommand*{\subsectionautorefname}{sec.}
\newcommand{\algorithmautorefname}{alg.}

\twocolumn[{%
\renewcommand\twocolumn[1][]{#1}%
\maketitle
\begin{center}
    \centering
    \includegraphics[width=\textwidth]{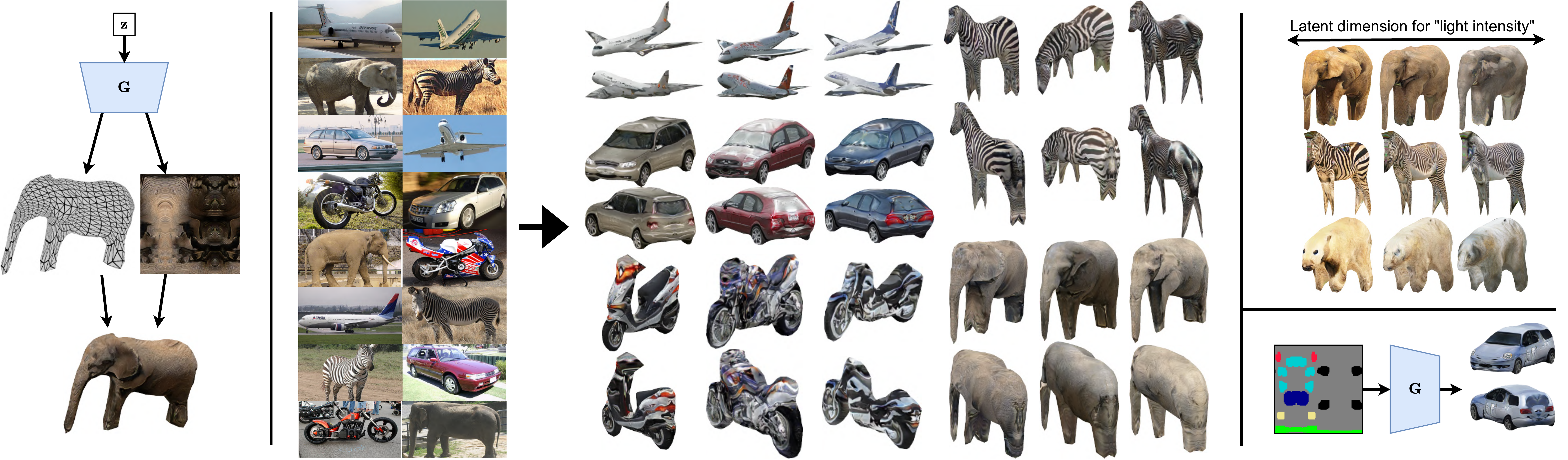}
    \captionof{figure}{\textbf{Left:} we focus on GANs, where our generator outputs a \emph{triangle mesh} and a UV-mapped texture. \textbf{Middle:} our method learns to synthesize textured 3D meshes given a real-world collection of 2D images.  \textbf{Top-right:} we showcase a setting where we train a single model to generate all classes. This model successfully disentangles some factors of the 3D environment (e.g.\ lighting/shadows) without explicit supervision. \textbf{Bottom-right:} we also demonstrate a conditional model that generates meshes from 3D semantic layouts.}
    \label{fig:teaser}
    \vspace{3mm}
\end{center}%
}]

\ificcvfinal\thispagestyle{empty}\fi

\begin{abstract}
Recent advances in differentiable rendering have sparked an interest in learning generative models of textured 3D meshes from image collections. These models natively disentangle pose and appearance, enable downstream applications in computer graphics, and improve the ability of generative models to understand the concept of image formation. Although there has been prior work on learning such models from collections of 2D images, these approaches require a delicate pose estimation step that exploits annotated keypoints, thereby restricting their applicability to a few specific datasets.
In this work, we propose a GAN framework for generating textured triangle meshes without relying on such annotations. We show that the performance of our approach is on par with prior work that relies on ground-truth keypoints, and more importantly, we demonstrate the generality of our method by setting new baselines on a larger set of categories from ImageNet -- for which keypoints are not available -- without any class-specific hyperparameter tuning. We release our code at {\footnotesize \url{https://github.com/dariopavllo/textured-3d-gan}}
\end{abstract}

\section{Introduction}
\label{sec:introduction}

Most of the recent literature in the field of generative models focuses on 2D image generation \cite{miyato2018cgans, zhang2018sagan, karras2019stylegan, brock2018biggan, karras2020styleganv2}, which largely ignores the fact that real-world images depict 2D projections of 3D objects. 
Constructing 3D generative models presents multiple advantages, including a fully disentangled control over shape, appearance, pose, as well as an explicit representation of spatial phenomena such as occlusions. While the controllability aspect of 2D generative models can be improved to some extent by disentangling factors of variation during the generation process~\cite{yang2017lrgan, singh2018finegan, karras2017progressivegan, karras2019stylegan}, the assumptions made by these approaches have been shown to be unrealistic without an inductive bias \cite{locatello2018challenging}. It is thus unclear whether 2D architectures can reach the same degree of controllability as a native 3D representation.\looseness=-1

As a result, a growing line of research investigates learning textured 3D mesh generators in both GAN \cite{pavllo2020convmesh, chen2019dibr} and variational settings \cite{henderson2020leveraging}. These approaches are trained with 2D supervision from a collection of 2D images, but require camera poses to be known in advance as learning a joint distribution over shapes, textures, and cameras is particularly difficult. Usually, the required camera poses are estimated from keypoint annotations using a factorization algorithm such as \emph{structure-from-motion} (SfM) \cite{marques2009estimating}. These keypoint annotations are, however, very expensive to obtain and are usually only available on a few datasets. %

In this work, we propose a new approach for learning generative models of textured triangle meshes with minimal data assumptions. Most notably, we do not require keypoint annotations, which are often not available in real-world datasets. Instead, we solely rely on: \emph{(i)} a single mesh template (optionally, a set of templates) for each image category, which is used to bootstrap the pose estimation process, and \emph{(ii)} a pretrained semi-supervised object detector, which we modify to infer semantic part segmentations on 2D images. These, in turn, are used to augment the initial mesh templates with a 3D semantic layout that allows us to refine pose estimates and resolve potential ambiguities.

First, we evaluate our approach on benchmark datasets for this task (Pascal3D+ \cite{liu2018beyond} and CUB \cite{wah2011cub}), for which keypoints are available, and show that our approach is quantitatively on par with the state-of-the-art \cite{pavllo2020convmesh} as demonstrated by FID metrics \cite{heusel2017ttur}, even though we do not use keypoints.
Secondly, we train a 3D generative model on a larger set of categories from ImageNet \cite{deng2009imagenet}, where we set new baselines without any class-specific hyperparameter tuning. To our knowledge, no prior works have so far succeeded in training textured mesh generators on real-world datasets, as they focus either on synthetic data or on simple datasets where poses/keypoints are available. %
We also show that we can learn a \emph{single} generator for all classes (as opposed to different models for each class, as done in previous work \cite{pavllo2020convmesh, chen2019dibr, henderson2020leveraging}) and notice the emergence of interesting disentanglement properties (e.g.\ color, lighting, style), similar to what is observed on large-scale 2D image generators \cite{brock2018biggan}.%

Finally, we quantitatively evaluate the pose estimation performance of our method under varying assumptions (one or more mesh templates; with or without semantic information), and showcase a proof-of-concept where 3D meshes are generated from sketches of semantic maps (\emph{semantic mesh generation}), following the paradigm of image-to-image translation. In summary, our main contributions are as follows:
\begin{itemize}[leftmargin=*, itemsep=-2pt]
    \item We introduce a new approach to 3D mesh generation that does not require keypoint annotations, enabling its use on a wider range of datasets as well as new image categories.
    \item We showcase 3D generative models in novel settings, including learning a \emph{single} 3D generator for all categories, and \emph{conditional generation} from semantic mesh layouts. In addition, we provide a preliminary analysis of the disentanglement properties learned by these models.
    \item We propose a comprehensive 3D pose estimation framework that combines the merits of template-based approaches and semantic-based approaches. We further extend this framework by explicitly resolving pose ambiguities and by adding support to multiple templates.
\end{itemize}

\section{Related work}
\label{sec:related-work}

\paragraph{Differentiable 3D representations} Recent work in 3D deep learning has focused on a variety of 3D representations. Among \emph{reconstruction} approaches, where the goal is to reconstruct 3D meshes from various input representations, \cite{park2019deepsdf} predict signed distance fields from point clouds, \cite{choy20163d, hane2017hierarchical, yang20173d, girdhar2016learning, zhu2017rethinking, wu2017marrnet, tatarchenko2017octree} predict 3D meshes from images using a voxel representation, and \cite{fan2017point} predict point clouds from images. These approaches require some form of 3D supervision, which is only achievable through synthetic datasets. More recent efforts have therefore focused on reconstructing meshes using 2D supervision from multiple views, e.g. \cite{yan2016perspective, gwak2017weakly, tulsiani2017multi, BMVC2017_99, tulsiani2018multi, yang2018learning} in the voxel setting, and \cite{insafutdinov2018unsupervised} using point clouds. However, the multiple-viewpoint assumption is unrealistic on real-world collections of natural images, which has motivated a new class of methods that aim to reconstruct 3D meshes from single-view images. Among recent works, \cite{kato2018n3mr, liu2019softras, kanazawa2018cmr, chen2019dibr, goel2020ucmr, li2020umr} are all based on this setting and adopt a \emph{triangle mesh} representation. Our work also focuses on triangle meshes due to their convenient properties: \emph{(i)} their widespread use in computer graphics, movies, video games; \emph{(ii)} their support for UV texture mapping, which decouples shape and color; \emph{(iii)} the ability of efficiently manipulating and transforming vertices via linear algebra. The use of triangle meshes in deep learning was recently enabled by \emph{differentiable renderers} \cite{loper2014opendr, kato2018n3mr, liu2019softras, chen2019dibr}, i.e.\ renderers that provide gradients w.r.t.\ scene parameters. Motivated by its support for UV maps, we use \emph{DIB-R} \cite{chen2019dibr} as our renderer of choice throughout this work.%

\paragraph{Keypoint-free pose estimation} The use of keypoints for pose estimation is limiting due to the lack of publicly available data and an expensive annotation process. Thus, a growing line of research focuses on inferring poses via semi-supervised objectives. 
To our knowledge, no approach has so far focused on \emph{generation}, but there have been some successful attempts in the \emph{reconstruction} literature. The initial pose estimation step of our framework is most closely related to \cite{goel2020ucmr, li2020umr}, which both propose approaches for 3D mesh \emph{reconstruction} without keypoints. In terms of assumptions, \cite{goel2020ucmr} require a canonical mesh template for each category. Object poses are estimated by fitting the mesh template to the silhouette of the object and by concurrently optimizing multiple camera hypotheses (which helps to deal with the large amount of bad local minima). \cite{li2020umr} do not require a mesh template, but instead use object part segmentations from a self-supervised model (\emph{SCOPS} \cite{hung2019scops}) to infer a 3D semantic template that is matched to the reference segmented image. Based on early experiments, we were unable to individually generalize these methods to \emph{generation} (our goal), which we found to have a lower tolerance to errors due to the intrinsic difficulty in training GANs. Instead, we here successfully combine both ideas (mesh templates \emph{and} semantics) and extend the overall framework with \emph{(i)} the optional support for multiple mesh templates, \emph{(ii)} a principled ambiguity resolution step that leverages part semantics to resolve conflicts among camera hypotheses with similar reprojection errors. We additionally adopt a more general object-part segmentation framework. Namely, we use a pre-trained semi-supervised object detector \cite{hu2018segmenteverything} modified to produce fine-grained semantic templates (\autoref{fig:dataset-demos}), as opposed to \emph{SCOPS} (used in \cite{li2020umr}), which we found to require class-specific hyperparameter tuning.%

\paragraph{Mesh generation} In the \emph{generation} literature there has been work on voxel representations \cite{wu2016learning,girdhar2016learning, smith2017improved, xie2018learning, zhu2018visual, balashova2018structure} and point clouds \cite{achlioptas2018learning, gadelha2018multiresolution}. These approaches require 3D supervision from synthetic data and are thus subject to the same limitations mentioned earlier. To our knowledge, the only approaches that tackle this task on a \emph{triangle mesh} setting using exclusively 2D supervision are \cite{henderson2020leveraging}, which focuses on a VAE setting using face colors (as opposed to full texture mapping) and is thus complementary to our work, and \cite{chen2019dibr, pavllo2020convmesh}, which adopts a GAN setting. In particular, \cite{chen2019dibr} represents the earliest attempt in generating textured 3D meshes using GANs, but their approach cannot supervise textures directly from image pixels. By contrast, the more recent \cite{pavllo2020convmesh} proposes a more comprehensive framework that can model both meshes and UV-mapped textures, which allows for successful application to natural images (albeit with keypoint annotations). We build upon \cite{pavllo2020convmesh}, from which we borrow the GAN architecture but substantially rework the supervision strategy to relax the keypoint requirement.%

\section{Method}
\label{sec:method}

\paragraph{Data requirements} As usual in both the \emph{reconstruction} \cite{kato2018n3mr, kanazawa2018cmr, goel2020ucmr, li2020umr} and \emph{generation} \cite{henderson2019learning, chen2019dibr, pavllo2020convmesh} literature, we require a dataset of segmented images. Segmentation masks (a.k.a.~\emph{silhouettes}) can easily be obtained through an off-the-shelf model (we use PointRend \cite{kirillov2020pointrend} pretrained on COCO \cite{lin2014mscoco}; details in Appendix \ref{sec:appendix-implementation-details}). Whereas prior approaches require keypoint annotations for every image, we only require an untextured mesh template for each image category, which can be downloaded freely from the web.
Optionally, our framework supports multiple mesh templates per category, a choice we explicitly evaluate in \autoref{sec:results}. We note that pose estimation from silhouettes alone can in some cases be ambiguous, and therefore we rely on object part semantics to resolve these ambiguities wherever possible. To this end, we use the semi-supervised, large-vocabulary object detector from \cite{hu2018segmenteverything, pavllo2020stylesemantics} to infer part segmentations on all images. We adopt their pretrained model as-is, without further training or fine-tuning, but post-process its output as described in Appendix \ref{sec:appendix-implementation-details}.%

\begin{figure}[t]
\begin{center}
  \includegraphics[width=\linewidth]{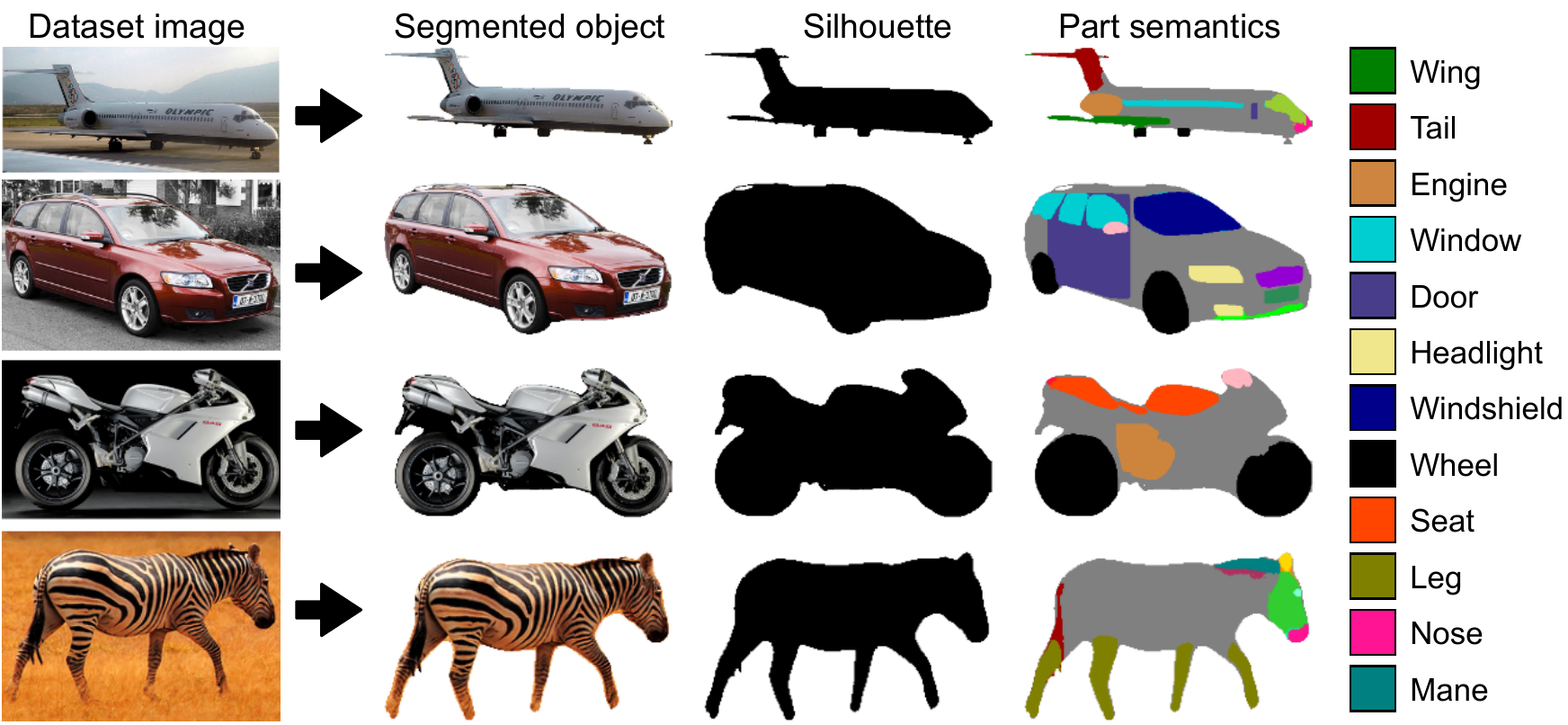}
\end{center}
   \vspace{-3mm}
   \caption{The dataset is initially processed into a clean collection of images with associated object masks and semantic part segmentations. This is done via off-the-shelf models and does not involve any additional data collection. Semantic classes have a precise meaning and are shared between different categories (e.g.\ wheels appear in both cars and motorbikes).}
\label{fig:dataset-demos}
\vspace{-4mm}
\end{figure}

\paragraph{Dataset preparation} Since our goal is to apply our method to real-world data that has not been manually cleaned or annotated -- unlike the commonly-used datasets CUB \cite{wah2011cub} and Pascal3D+ \cite{liu2018beyond} -- we attempt to automatically detect and remove images that do not satisfy some quality criteria. In particular, objects should not be \emph{(i)} too small, \emph{(ii)} truncated, or \emph{(iii)} occluded by other objects (implementation details in Appendix \ref{sec:appendix-implementation-details}). This filtering step is tuned for high precision and low recall, as we empirically found that it is beneficial to give more importance to the former. All our experiments and evaluations (\autoref{sec:experiments}) are performed on the dataset that results from this step.
Finally, sample images and corresponding silhouettes/part segmentations can be seen in \autoref{fig:dataset-demos}, which also highlights how some semantic parts are shared across image categories.

\begin{figure*}[t]
    \begin{center}
      \includegraphics[width=\textwidth]{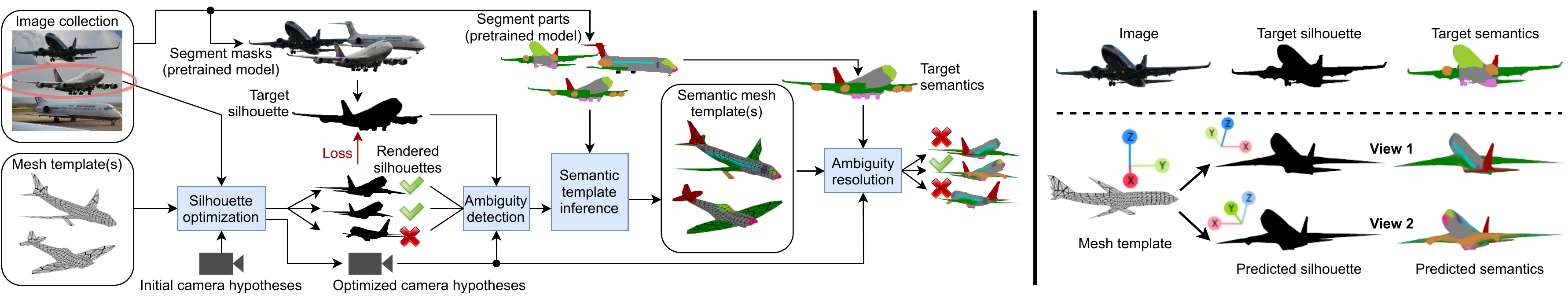}
    \end{center}
    \vspace{-4mm}
    \caption{\textbf{Left:} schematic overview of the proposed pose estimation pipeline. The left side shows our data requirements (a collection of 2D images and one or more untextured mesh templates). For clarity, we only show the optimization process for the circled airplane, although the \emph{semantic template inference} step involves multiple instances. \textbf{Right:} ambiguity arising from opposite poses. The two camera hypotheses produce almost-identical silhouettes which closely approximate the target, but describe opposite viewpoints. This particular example would initially be rejected by our ambiguity detection test, but it would then be resolved once semantics are available.}
\label{fig:pipeline-overview}
\vspace{-3mm}
\end{figure*}

\subsection{Pose estimation framework}
\label{sec:pose-estimation}

\paragraph{Overview} Most \emph{reconstruction} and \emph{generation} approaches require some form of pose estimation to initialize the learning process. Jointly learning a distribution over camera poses and shapes/textures is extremely challenging and might return a trivial solution that does not entail any 3D reasoning. Therefore, our approach also requires a pose estimation step in order to allow the learning process to converge to meaningful solutions. Our proposed pose estimation pipeline is summarized in \autoref{fig:pipeline-overview}: starting from a set of randomly-initialized camera hypotheses for each object instance, we render the mesh template(s) using a differentiable renderer and optimize the camera parameters so that the rendered silhouette matches the target silhouette of the object. At this point, no semantics, colors, or textures are involved, so the approach can lead to naturally ambiguous poses (see \autoref{fig:pipeline-overview} right, for an example). We then introduce a novel ambiguity detection step to select only images whose inferred pose is unambiguous, and use the most confident ones to infer a 3D \emph{semantic template}, effectively augmenting the initial mesh templates with semantic information (more examples of such templates can be seen in \autoref{fig:semantic-templates}). Afterwards, the process is repeated -- this time leveraging semantic information -- to resolve ambiguities and possibly reinstate images that were previously discarded. The final output is a camera pose for each object as well as a confidence score that can be used to trade off recall (number of available images) for precision (similarity to ground-truth poses). In the following, we describe each step in detail.

\paragraph{Silhouette optimization} The first step is a fitting procedure applied separately to each image. Following \cite{goel2020ucmr}, who observe that optimizing multiple camera hypotheses with differing initializations is necessary to avoid local minima, we initialize a set of $N_c$ camera hypotheses for each image as described in Appendix \ref{sec:appendix-implementation-details}. 
Our camera projection model is the \emph{augmented} weak-perspective model of \cite{pavllo2020convmesh}, which comprises a rotation $\mathbf{q} \in \mathbb{R}^4$ (a unit quaternion), a scale $s \in \mathbb{R}$, a screen-space translation $\mathbf{t} \in \mathbb{R}^2$, and a perspective correction term $z_0 \in \mathbb{R}$ which is used to approximate perspective distortion for close objects. 
We minimize the mean squared error (MSE) in pixel space between the rendered silhouette $\mathcal{R(\cdot)}$ and the target silhouette $\mathbf{x}$:
\setlength{\abovedisplayskip}{3pt}
\setlength{\belowdisplayskip}{3pt}
\begin{equation}
    \min_{\mathbf{q}, \mathbf{t}, s, {z0}} \left\lVert \mathcal{R}(\mathbf{V}_\text{tpl}, \mathbf{F}_\text{tpl};\; \mathbf{q}, \mathbf{t}, s, {z_0}) - \mathbf{x} \right\rVert^2,
\end{equation}
where $\mathcal{R}$ is the differentiable rendering operation, $\mathbf{V}_\text{tpl}$ represents the (fixed) mesh template vertices, and $\mathbf{F}_\text{tpl}$ represents the mesh faces.
Each camera hypothesis is optimized 
using a variant of Adam \cite{kingma2014adam} that implements full-matrix preconditioning as opposed to a diagonal one. Given the small number of learnable parameters (8 for each hypothesis), the $\mathcal{O}(n^3)$ cost of inverting the preconditioning matrix is negligible compared to the convergence speed-up. We provide hyperparameters and more details about this choice in the Appendix \ref{sec:appendix-implementation-details}. 
In the settings where we use multiple mesh templates $N_t$, we simply replicate each initial camera hypothesis $N_t$ times so that the total number of hypotheses to optimize is $N_c \cdot N_t$. In this case, we compensate for the increase in optimization time by periodically pruning the worst camera hypotheses during optimization. Additionally, in all settings, we start by rendering at a low image resolution and progressively increase the resolution over time, which further speeds up the process. We describe how both strategies are implemented in the Appendix \ref{sec:appendix-implementation-details}.

\paragraph{Scoring and ambiguity detection} All symmetric objects (i.e.\ many natural and man-made objects) present \emph{ambiguous poses}: opposite viewpoints that produce the same silhouette after 2D projection (\autoref{fig:pipeline-overview} right).  %
Similar ambiguities can also arise as a result of noisy segmentation masks, inappropriate mesh templates, or camera hypotheses that converge to bad local minima.
Since wrong pose estimates have a significant negative impact on the rest of the pipeline, this motivates the design of an \emph{ambiguity detection} step. %
Ideally, we would like to accept pose estimates that are both \emph{confident} -- using the \emph{intersection-over-union} (IoU) between the rendered/target silhouettes as a proxy measure -- and \emph{unambiguous}, i.e.\ no two camera hypotheses with high IoU should describe significantly different poses.
We formalize this as follows: we first score each hypothesis $k$ as $(\mathbf{v}_\text{conf})_k = (\text{softmax}(\mathbf{v}_\text{IoU}\,/\,\tau))_k$, where $\tau = 0.01$ is a temperature coefficient that gives similar weights to IoU values that are close to the maximum, and low weights to IoU values that are significantly lower than the maximum. Next, we require that highly-confident poses (as measured by $\mathbf{v}_\text{conf}$) should describe similar rotations. We therefore construct a pairwise distance matrix $\mathbf{D}$ of shape $N_c \times N_c$, where each entry $d_{ij}$ describes the geodesic distance between the rotation of the $i$-th hypothesis and the rotation of the $j$-th hypothesis. Entries are then weighted by $\mathbf{v}_\text{conf}$ across both rows and columns, and are finally summed up, yielding a scalar agreement score $v_\text{agr}$ for each image:
\setlength{\abovedisplayskip}{3pt}
\setlength{\belowdisplayskip}{3pt}
\begin{equation}
    \mathbf{D} = 1 - (\mathbf{Q}^T \mathbf{Q})^{\circ 2}, \quad
    v_\text{agr} = \left\lVert \mathbf{D} \odot (\mathbf{v}_\text{conf}\, \mathbf{v}_\text{conf}^T)  \right\rVert_1
\end{equation}
where $\mathbf{Q}$ is a $4 \times N_c$ matrix of unit quaternions (one per hypothesis), $\mathbf{M}^{\circ 2}$ denotes the element-wise square, and $\odot$ denotes the element-wise product.

The agreement score $v_\text{agr}$ can be roughly interpreted as follows: a score of 0 (best) implies that all \emph{confident} camera hypotheses describe the same rotation (they agree with each other). A score of 0.5 describes two poses that are rotated by 180 degrees from one another\footnote{For example, consider a $\mathbf{D}$ matrix of size $2\times2$, where entries along the main diagonal are 0, and 1 elsewhere.}. Empirically, we established that images with $v_\text{agr} > 0.3$ should be rejected.

\paragraph{Semantic template inference} Simply discarding ambiguous images might significantly reduce the size and diversity of the training set. Instead, we propose to resolve the ambiguous cases.
While this is hardly possible when we only have access to silhouettes, it becomes almost trivial once \emph{semantics} are available (\autoref{fig:pipeline-overview} right). 
A similar idea was proposed in \cite{li2020umr}, who infer a 3D semantic template by averaging instances that are close to a predetermined exemplar (usually an object observed from the left or right side). Yet, our formulation does not require an exemplar but directly leverages samples that have passed the ambiguity detection test. Since our data requirements assume that mesh templates are untextured, our first step in this regard aims at augmenting each mesh template with part semantics. Among images that have passed the ambiguity test ($v_\text{agr} < 0.3$), we select the camera hypothesis with the highest IoU. For each mesh template, the semantic template is computed using the top $N_\text{top}=100$ images assigned to that template, as measured by the IoU.
Then, we frame this step as an optimization problem where the goal is to learn vertex colors while keeping the camera poses fixed, minimizing the MSE between the rendered (colored) mesh template and the 2D image semantics, averaged among the top samples:
\setlength{\abovedisplayskip}{3pt}
\setlength{\belowdisplayskip}{3pt}
\begin{equation}
\resizebox{0.9\linewidth}{!}{$\displaystyle\min_{\mathbf{C}_\text{tpl}} \frac{1}{N_\text{top}} \sum_i \left\lVert \mathcal{R}(\mathbf{V}_\text{tpl}, \mathbf{F}_\text{tpl}, \mathbf{C}_\text{tpl};\; \mathbf{q_i}, \mathbf{t_i}, s_i, {z_0}_i) - \mathbf{C_i} \right\rVert^2 ,$}
\end{equation}
where $\mathbf{C}_\text{tpl}$ represents the vertex colors of the template and $\mathbf{C_i}$ denotes the 2D semantic image. For convenience, we represent $\mathbf{C}_\text{tpl}$ as a $K \times N_{v}$ matrix, where $N_{v}$ is the number of vertices and $K$ is the number of semantic classes (color channels, not necessarily limited to 3), and $\mathbf{C_i}$ is a $K \times N_{pix}$ matrix, where $N_{pix}$ is the number of image pixels. %
In the Appendix \ref{sec:appendix-implementation-details}, we derive an efficient closed-form solution that requires only a single pass through the dataset. Examples of the resulting semantic templates are shown in \autoref{fig:semantic-templates}.

\paragraph{Ambiguity resolution} In the last step of our pose estimation pipeline, we repeat the scoring process described in ``\emph{Scoring and ambiguity detection}" with the purpose of resolving ambiguities. Instead of evaluating the scores on the IoU, however, we use the mean intersection-over-union (mIoU) averaged across semantic classes. Since our inferred semantic templates are continuous, we adopt a smooth generalization of the mIoU (weighted Jaccard similarity) in place of the discrete version:
\setlength{\abovedisplayskip}{3pt}
\setlength{\belowdisplayskip}{3pt}
\begin{equation}
    \resizebox{0.5\linewidth}{!}{$\displaystyle\text{mIoU} =  \frac{1}{K}\sum_k \frac{\lVert \min(\mathbf{\hat{C}_k}, \mathbf{C_k}) \rVert_1}{\lVert \max(\mathbf{\hat{C}_k}, \mathbf{C_k}) \rVert_1}$}
\end{equation}
where $\mathbf{\hat{C}_k}$ is the rendered semantic class $k$ and $\min, \max$ (performed element-wise) represent the weighted intersection and union, respectively.
We then recompute the confidence scores and agreement scores as before (using the mIoU as a target metric), discard the worst 10\% images in terms of mIoU as well as those whose $v_\text{agr} > 0.3$, and select the best hypothesis for each image as measured by the mIoU. We found no practical advantage in repeating the semantic template inference another time, nor in re-optimizing/fine-tuning the camera poses using semantics. We show this quantitatively in \autoref{sec:results} and discuss further details on various exploratory attempts in Appendix \ref{sec:appendix-negative-results}.\looseness=-1

\begin{figure}[t]
\begin{center}
  \includegraphics[width=\linewidth]{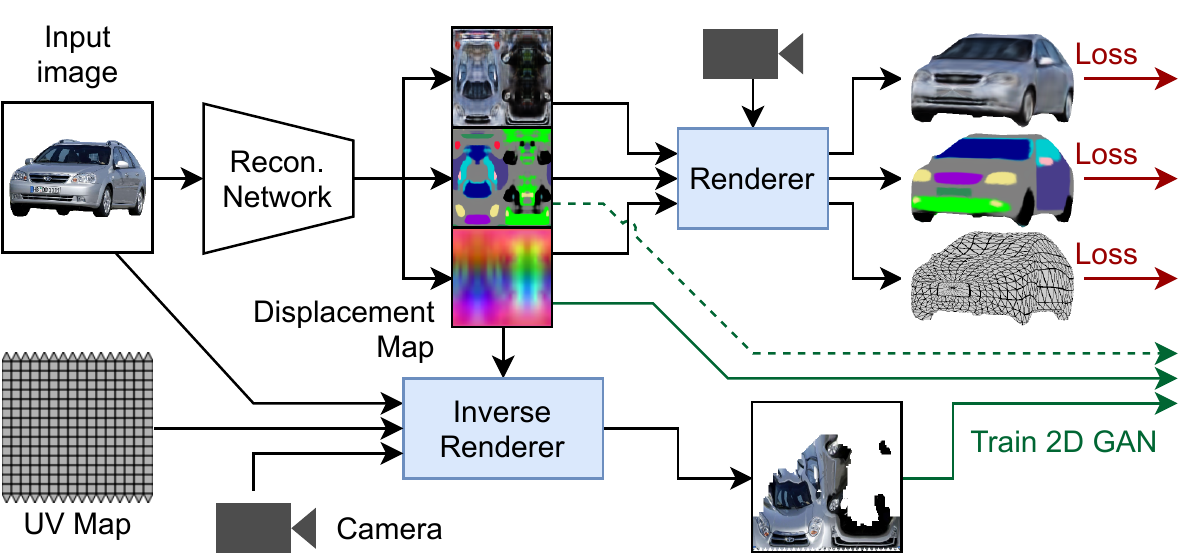}
\end{center}
\vspace{-3mm}
   \caption{Generation framework using the \emph{convolutional mesh} representation. Images are fed into a network trained to reconstruct meshes (parameterized as 2D displacement maps), given camera poses. The meshes are then used to project natural images onto the UV map. Finally, the resulting partial textures, displacement maps, and (optionally) predicted semantics are used to train a 2D convolutional GAN in UV space.}
\label{fig:generation-framework}
\vspace{-4mm}
\end{figure}

\subsection{Generation framework}
\label{sec:generation-estimation}

\begin{figure*}[t]
\begin{center}
  \includegraphics[width=\textwidth]{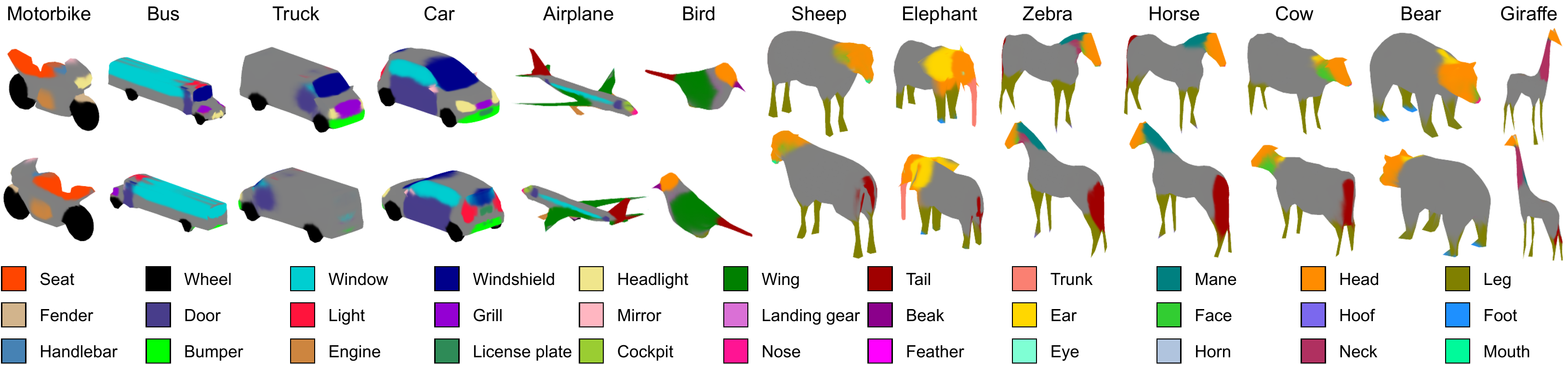}
\end{center}
\vspace{-2mm}
   \caption{Learned 3D semantic templates. We show one template per category from two views (front/back). Colors are exaggerated for presentation purposes, but in practice the probability maps are smoother. We also highlight how semantic parts are shared among categories.}
\label{fig:semantic-templates}
\vspace{-4mm}
\end{figure*}

The camera poses obtained using the approach described in \autoref{sec:pose-estimation} can be used to train a generative model as shown in \autoref{fig:generation-framework}. For this component, we build upon \cite{pavllo2020convmesh}, from which we borrow the \emph{convolutional mesh} representation and the GAN architecture.
Our generation approach mainly consists of three steps. \emph{(i)} Given a collection of images, segmentation masks, and their poses\footnote{In \cite{pavllo2020convmesh}, poses are estimated via structure-from-motion on ground-truth keypoints. In this work, we use our proposed approach (\autoref{sec:pose-estimation}).}, we train a reconstruction model to predict mesh, texture, and semantics given only the 2D image as input. Although predicted textures are not used in subsequent steps (the GAN learns directly from image pixels), \cite{pavllo2020convmesh} observe that predicting textures during training has a beneficial regularizing effect on the mesh, and therefore we also keep this reconstruction term. Unlike \cite{pavllo2020convmesh} (where semantics were not available), however, we also predict a 3D semantic part segmentation in UV space, which provides further regularization and enables interesting conditional generation settings (we showcase this in \autoref{sec:results}).
As in \cite{pavllo2020convmesh}, we parameterize the mesh as a 2D displacement map that deforms a sphere template in its tangent space.
\emph{(ii)} Through an inverse rendering approach, image pixels are projected onto the UV map of the mesh, yielding partially-occluded textures. Occlusions are represented as a binary mask in UV space. \emph{(iii)} Finally, displacement maps and textures are modeled in UV space using a standard 2D convolutional GAN, whose training strategy compensates for occlusions by masking inputs to the discriminator.

\paragraph{Architecture} Our experiments (\autoref{sec:experiments}) analyze two different settings:  \textbf{A} where we train a separate model for each category, and \textbf{B} where we train a single model for all categories. 
In setting \textbf{A}, we reuse similar reconstruction and GAN architectures to~\cite{pavllo2020convmesh} in order to establish a fair comparison with their approach.
We only modify the output head of the reconstruction model, where we add $K$ extra output channels for the semantic class prediction ($K$ depends on the category).
In setting \textbf{B}, we condition the model on the object category by modifying all Batch\-Norm layers and learning different gain and bias parameters for each category. Additionally, in the output head we share semantic classes among categories (for instance there is a unique output channel for \emph{wheel} that is shared for buses, trucks, \emph{etc.}; see \autoref{fig:semantic-templates}). We do not make any other change that would affect the model's capacity. As for the GAN, in both \textbf{A} and \textbf{B}, we use the same architecture as \cite{pavllo2020convmesh}. Further details regarding hyperparameters, implementation and optimizations to improve rendering speed can be found in Appendix \ref{sec:appendix-implementation-details}.%

\paragraph{Loss} The reconstruction model is trained to jointly minimize the MSE between \emph{(i)} rendered and target silhouettes, \emph{(ii)} predicted RGB texture and target 2D image, \emph{(iii)} predicted semantic texture (with $K$ channels) and target 2D semantic image. As in \cite{pavllo2020convmesh}, we add a smoothness loss to encourage neighboring faces to have similar normals. Finally, the availability of mesh templates allows us to incorporate a strong shape prior into the model via a loss term that can be regarded as an extreme form of semi-supervision: on images with very confident poses (high IoU), we provide supervision directly on the predicted 3D vertices by adding a MSE loss between the latter and the vertices of the mesh template (i.e.\ our surrogate ground-truth), only on the top 10\% of images as measured by the IoU. This speeds up convergence and helps with modeling fine details such as wings of airplanes, where silhouettes alone provide a weak learning signal from certain views. This step requires \emph{remeshing} the templates to align them to a common topology, which we describe in Appendix \ref{sec:appendix-implementation-details}.

\begin{figure*}[t]
\begin{center}
  \includegraphics[width=\textwidth]{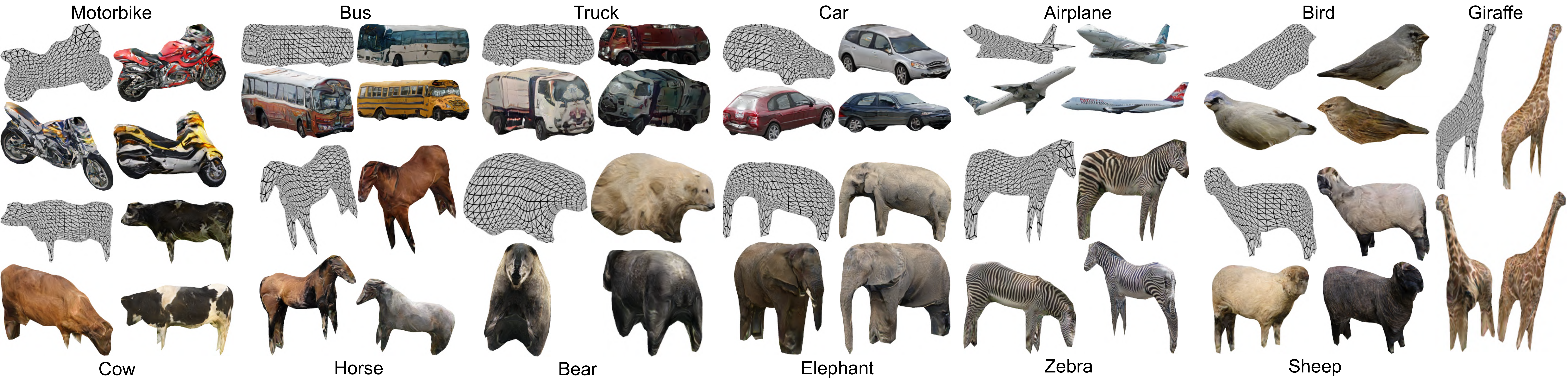}
\end{center}
\vspace{-4mm}
   \caption{Qualitative results for all 13 classes used in our work. For each class, we show one wireframe mesh on the left, the corresponding textured mesh on the right, and two additional textured meshes on the second row. Meshes are rendered from random viewpoints.}
\label{fig:qualitative-results}
\vspace{-4mm}
\end{figure*}

\vspace{-2mm}
\section{Experiments}
\label{sec:experiments}
\vspace{-1mm}
We quantitatively evaluate the aspects that are most central to our approach: pose estimation and generation quality.

\paragraph{Pose estimation} On datasets where annotated keypoints are available, we compare the poses estimated by our approach to poses estimated from \emph{structure-from-motion} (SfM) on ground-truth keypoints. Since the robustness of SfM depends on the number of visible keypoints, we never refer to SfM poses as ``ground-truth poses", as these are not available in the real-world datasets we use. Nonetheless, we believe that SfM poses serve as a good approximation of ground-truth poses on most images. Our evaluation metrics comprise \emph{(i)} the \emph{geodesic distance} (GD) between the rotation $\mathbf{q}$ predicted by our approach and the SfM rotation $\mathbf{p}$, defined as $\text{GD} = 1 - (\mathbf{p} \cdot \mathbf{q})^2$ for quaternions, where $\text{GD} \in [0, 1]$ \footnote{More commonly known as \emph{cosine distance} when quaternions are used to describe orientations, as in our case.}; and \emph{(ii)} the \emph{recall}, which measures the fraction of usable images that have passed the ambiguity detection test.
We evaluate pose estimation at different stages: after silhouette optimization (where no semantics are involved), and after the semantic template inference. Additionally, we compare settings where only one mesh template per category is available, and where multiple mesh templates are employed (we use 2--4 templates per category).

\paragraph{Generative modeling} Following prior work on textured 3D mesh generation with GANs \cite{pavllo2020convmesh}, we evaluate the Fréchet Inception Distance (FID) \cite{heusel2017ttur} on meshes rendered from random viewpoints. For consistency, our implementation of this metric follows that of \cite{pavllo2020convmesh}. Since our pose estimation framework discards ambiguous images and the FID is sensitive to the number of evaluated images, we \emph{always} use the full dataset for computing reference statistics. As such, there is an incentive for optimizing both \emph{GD} and \emph{recall} metrics as opposed to trading one off for the other. Finally, consistently with \cite{pavllo2020convmesh}, we generate displacement maps at $32 \times 32$ resolution, textures at $512 \times 512$, and sample from the generator using a truncated Gaussian at $\sigma = 1.0$.

\subsection{Datasets}
\label{sec:datasets}
We evaluate our approach on three datasets: CUB-200-2011  (CUB) \cite{wah2011cub}, Pascal3D+ (P3D) \cite{liu2018beyond}, and a variety of classes from ImageNet \cite{deng2009imagenet}. The first two provide keypoint annotations and serve as a comparison to previous work, whereas on the latter we set new baselines. Combining all datasets, we evaluate our approach on 13 categories.

\paragraph{CUB (Birds)} For consistency with prior work, we adopt the split of \cite{pavllo2020convmesh, kanazawa2018cmr} ($\approx$6k training images). As we work in the unconditional setting, we do not use class labels.

\paragraph{Pascal3D+ (P3D)} Again, we adopt the split of \cite{pavllo2020convmesh, kanazawa2018cmr}, and test our approach on both \emph{car} and \emph{airplane} categories. Since \cite{pavllo2020convmesh} has only tested on cars, we train the model of \cite{pavllo2020convmesh} on airplanes and provide a comparison here. P3D comprises a subset of images from ImageNet and \cite{pavllo2020convmesh} evaluates only on this subset; for consistency, we adopt the same strategy.

\paragraph{ImageNet} Our final selection of classes 
comprises the vehicles and animals that can be seen in \autoref{fig:semantic-templates}/\ref{fig:qualitative-results}. The list of \emph{synsets} used in each class as well as summary statistics are provided in the Appendix \ref{sec:appendix-dataset-information}. The set of ImageNet classes includes \emph{car} and \emph{airplane}, which partially overlap with P3D. Therefore, when we mention these two classes, we always specify the subset we refer to (ImageNet or P3D). We also note that the dataset is heavily imbalanced, ranging from $\approx$300 usable images for \emph{giraffe} to thousands of images for \emph{car}. For this reason, in setting \textbf{B} we take measures to balance the dataset during training (Appendix \ref{sec:appendix-implementation-details}). 

\begin{table}[t]
\begin{center}
\resizebox{\linewidth}{!}{
\setlength{\tabcolsep}{2pt}
\renewcommand{\arraystretch}{1.1}
\begin{tabular}{cc||c|rl||c|rl||c|rl}
\multicolumn{1}{l}{}                                                                               &             & \multicolumn{3}{c||}{Bird}                         & \multicolumn{3}{c||}{Car}                          & \multicolumn{3}{c}{Airplane}                     \\
\multicolumn{1}{c|}{Setting}                                                                       & Step        & GD(1)         & \multicolumn{2}{c||}{GD (Recall)} & GD(1)         & \multicolumn{2}{c||}{GD (Recall)} & GD(1)         & \multicolumn{2}{c}{GD (Recall)} \\ \hline
\multicolumn{1}{c|}{\multirow{3}{*}{\begin{tabular}[c]{@{}c@{}}Single\\ template\end{tabular}}}    & Silhouette & 0.47          & 0.35                  & (52\%)   & 0.12          & {\ul 0.05}         & (75\%)      & 0.31          & 0.28               & (85\%)      \\
\multicolumn{1}{c|}{}                                                                              & Semantics  & \textbf{0.29} & \textbf{0.24}         & (74\%)   & 0.11          & 0.06               & (84\%)      & 0.25          & 0.18               & (78\%)      \\
\multicolumn{1}{c|}{}                                                                              & Repeat x2  & \textbf{0.29} & {\ul \textbf{0.24}}   & (76\%)   & 0.15          & 0.11               & (85\%)      & 0.24          & 0.17               & (75\%)      \\ \hline
\multicolumn{1}{c|}{\multirow{3}{*}{\begin{tabular}[c]{@{}c@{}}Multiple\\ templates\end{tabular}}} & Silhouette & 0.47          & 0.33                  & (44\%)   & 0.10          & {\ul 0.05}         & (78\%)      & 0.28          & 0.22               & (81\%)      \\
\multicolumn{1}{c|}{}                                                                              & Semantics  & {\ul 0.32}    & {\ul 0.27}            & (76\%)   & \textbf{0.06} & \textbf{0.04}      & (88\%)      & {\ul 0.22}    & \textbf{0.15}      & (79\%)      \\
\multicolumn{1}{c|}{}                                                                              & Repeat x2  & {\ul 0.32}    & {\ul 0.27}            & (78\%)   & {\ul 0.07}    & {\ul 0.05}         & (89\%)      & \textbf{0.21} & {\ul 0.16}         & (80\%)     
\end{tabular}
}
\end{center}
\vspace{-2mm}
\caption{Pose estimation results under different settings. Best in \textbf{bold}; second best {\ul underlined}. We report \emph{geodesic distance} (GD; lower = better) after each step and associated recall (higher = better) arising from ambiguity detection.
For comparison, we also report GD w/o ambiguity detection, GD(1), assuming 100\% recall.\looseness=-1}
\label{tab:pose-estimation-results}
\vspace{-4mm}
\end{table}

\subsection{Results}
\label{sec:results}

\begin{figure}[b]
\vspace{-6mm}
\begin{center}
  \includegraphics[width=\linewidth]{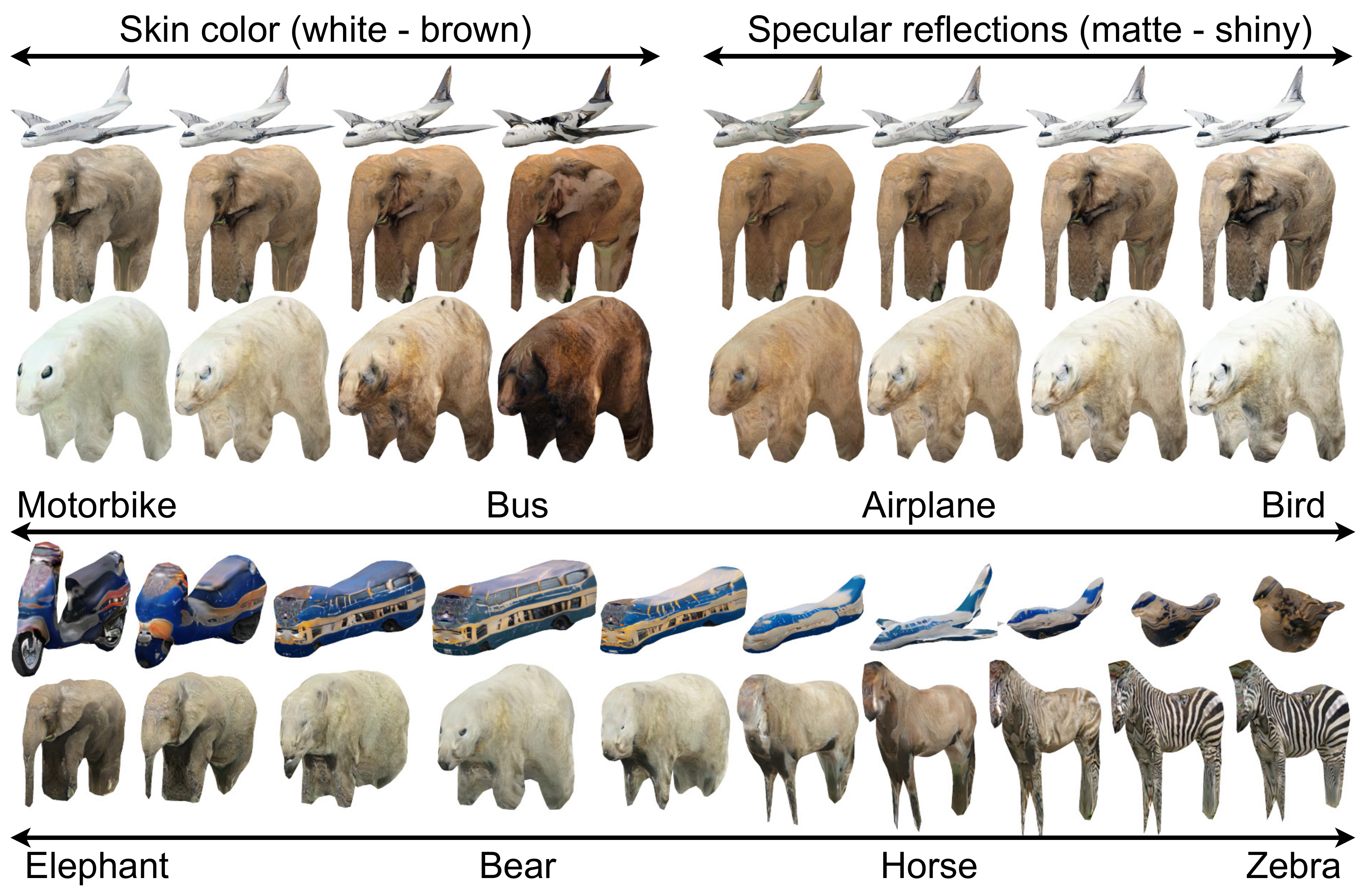}
\end{center}
\vspace{-4mm}
   \caption{Disentanglement and interpolation in the model trained to generate all classes (setting \textbf{B}). \textbf{Top:} directions in latent space that correlate with certain style factors, such as skin color and lighting. The effect is consistent across different classes. \textbf{Bottom:} interpolation between different classes with a fixed latent code.}
\label{fig:disentanglement-interpolation}
\end{figure}

\begin{table*}[t]
\begin{center}
\setlength{\tabcolsep}{2pt}
\renewcommand{\arraystretch}{1.1}
\resizebox{0.66\textwidth}{!}{
\begin{tabular}{l|lllllllllllll|l}
Setting                      & MBike      & Bus            & Truck          & Car            & Airplane       & Bird     & Sheep          & Elephant       & Zebra          & Horse          & Cow            & Bear           & Giraffe        & All            \\ \hline
Single TPL (\textbf{A}) & 107.4          & 219.3          & \textbf{164.1} & \textbf{30.73} & 77.84          & \textbf{55.75} & 173.7          & \textbf{114.5} & 28.19          & 113.3          & 137.0          & {\ul 187.1}    & {\ul 157.7}    & --             \\
Multi TPL (\textbf{A})  & 107.0          & \textbf{160.7} & 206.1          & {\ul 32.19}    & 102.2          & {\ul 56.54}    & \textbf{155.1} & 135.9          & \textbf{22.10} & {\ul 107.1}    & {\ul 133.0}    & 195.5          & \textbf{126.0} & --             \\ \hline
Single TPL (\textbf{B})       & {\ul 94.74}    & 204.98         & {\ul 179.3}    & 39.68          & \textbf{46.46} & 88.47          & 169.9          & {\ul 127.6}    & {\ul 24.47}    & \textbf{106.9} & 139.4          & \textbf{156.4} & 176.8          & \textbf{60.82} \\
Multi TPL (\textbf{B})        & \textbf{94.03} & {\ul 187.75}   & 204.7          & 46.11          & {\ul 77.27}    & 77.23          & {\ul 163.8}    & 146.2          & 31.70          & 113.4          & \textbf{117.5} & 189.9          & 158.0          & {\ul 63.00}   
\end{tabular}
}
\resizebox{0.33\textwidth}{!}{
\renewcommand{\arraystretch}{1.1}
\begin{tabular}{l|lll}
Method            & Bird (CUB)     & Car (P3D)      & Airplane (P3D) \\ \hline
Keypoints+SfM \cite{pavllo2020convmesh}  & \textbf{41.56} & 43.09          & 147.8*          \\ \hline
Silhouette (single TPL) & 73.67          & 38.16           & 100.5     \\
Silhouette (multi TPL)  & 88.39          & \textbf{36.17} & 96.28 \\ \hline
Semantics (single TPL) & {\ul 55.75}          & {\ul 36.52}           & \textbf{81.28}           \\
Semantics (multi TPL)  & 56.54          & 37.56 & {\ul 88.85}
\end{tabular}
}
\end{center}
\vspace{-2mm}
\caption{\textbf{Left:} FID of our approach on ImageNet (except \emph{bird}, which refers to CUB). We report results for models trained separately on different classes (setting \textbf{A}) and a single model that generates all classes (setting \textbf{B}). \textbf{Right:} comparison of our FID w.r.t.\ prior work, using either silhouettes alone or our full pipeline. * = trained by us; TPL = mesh template(s); lower = better, best in \textbf{bold}, second best {\ul underlined}.}
\label{tab:fid-scores-all}
\vspace{-5mm}
\end{table*}

\paragraph{Pose estimation} We evaluate our pose estimation framework on \emph{bird}, \emph{car}, and \emph{airplane}, for which we have keypoint annotations. Reference poses are obtained using the SfM implementation of \cite{kanazawa2018cmr}. For birds (CUB), the scores are computed on all images, whereas for cars/airplanes they are computed on the overlapping images between P3D and our ImageNet subset. Results are summarized in \autoref{tab:pose-estimation-results}. Interestingly, using multiple mesh templates does not seem to yield substantially different results, suggesting that our approach can work effectively with as little as one template per class. Moreover, incorporating semantic information improves both GD and recall. Finally, we repeat the ambiguity detection and semantic template inference steps a second time, but observe no improvement. Therefore, in our following experiments we only perform these steps once. We further discuss these results in Appendix \ref{sec:appendix-additional-results}, where we aim to understand the most common failure modes by analyzing the full distribution of rotation errors. Qualitatively, the inferred 3D semantic templates can be found in \autoref{fig:semantic-templates}.

\paragraph{Generative model} We report the FID on ImageNet in \autoref{tab:fid-scores-all}, left (\emph{bird} refers to CUB), where we set new baselines. As before, we compare settings where we adopt a single mesh template \emph{vs} multiple templates. We also showcase a conditional model that learns to synthesize all categories using a single generator (setting \textbf{B}). Although this model has the same capacity as the individual models (but was trained to generate all classes at once), we note that its scores are in line with those of setting \textbf{A}, and in some classes (e.g.\ \emph{airplane}) they are significantly better, most likely due to a beneficial regularizing effect. However, we also note that there is no clear winner on all categories.
To our knowledge, no prior work has trained a single 3D generator on multiple categories without some form of supervision from synthetic data. %
Therefore, in one of the following paragraphs we analyze this model from a disentanglement perspective.
Next, in \autoref{tab:fid-scores-all} (right), we compare our results to the state-of-the-art \cite{pavllo2020convmesh} on the \emph{bird}, \emph{car}, and \emph{airplane} categories from CUB/P3D. We find that our approach outperforms \cite{pavllo2020convmesh} on \emph{car} and \emph{airplane} (P3D) -- even though we do not exploit ground-truth keypoints -- and performs slightly worse on \emph{bird} (CUB). We speculate this is mainly due to the fact that, on CUB, all keypoints are annotated (including occluded ones), whereas P3D only comprises annotations for visible keypoints, potentially reducing the effectiveness of SfM as a pose estimation method. Finally, we point out that although there is a large variability among the scores across classes, comparing FIDs only makes sense within the same class, since the metric is affected by the number of images. %

\paragraph{Qualitative results} In addition to those presented in \autoref{fig:teaser}, we show further qualitative results in \autoref{fig:qualitative-results}. For animals, we observe that generated textures are generally accurate (e.g.\ the high-frequency details of zebra stripes are modeled correctly), with occasional failures to model facial details. With regards to shape, legs are sporadically merged but also appear correct on many examples. We believe these issues are mostly due to a pose misalignment, as animals are deformable but our mesh templates are rigid. As part of future work, we would like to add support for articulated mesh templates \cite{kulkarni2020acsm} to our method. 
As for vehicles, the generated shapes are overall faithful to what one would expect, especially on airplanes where modeling wings is very challenging. We also note, however, that the textures of rare classes (\emph{truck} above all) present some incoherent details. Since we generally observe that the categories with more data are also those with the best results, these issues could in principle be mitigated by adding more images. Finally, we show additional qualitative results in the Appendix \ref{sec:appendix-additional-results}.

\paragraph{Disentanglement and interpolation} We attempt to interpret the latent space of the model trained to synthesize all classes (setting \textbf{B}), following \cite{harkonen2020ganspace}. We identify some directions in the latent space that correlate with characteristics of the 3D scene, including light intensity (\autoref{fig:teaser}, top-right), specular reflections and color (\autoref{fig:disentanglement-interpolation}). Importantly, these factors seem to be shared across different classes and are learned without explicit supervision. Although our analysis is preliminary, our findings suggest that 3D GANs disentangle high-level features in an interpretable fashion, similar to what is observed in 2D GANs to some extent (e.g.\ on pose and style). However, since 3D representations already disentangle appearance and pose, the focus of the disentangled features is on other aspects such as texture and lighting.
\autoref{fig:disentanglement-interpolation} (bottom) illustrates interpolation between different classes while keeping the latent code fixed. Style is preserved and there are no observable artifacts, suggesting that the latent space is structured.

\begin{figure}[t]
\vspace{-2mm}
\begin{center}
  \includegraphics[width=\linewidth]{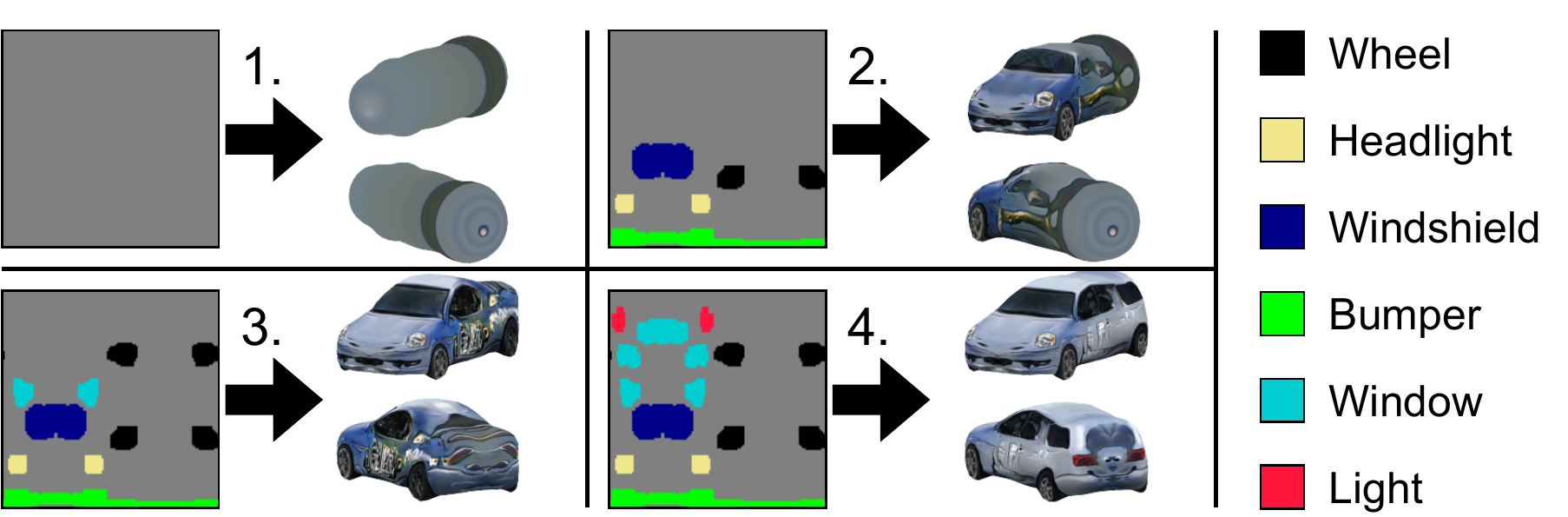}
\end{center}
\vspace{-4mm}
   \caption{Conditional mesh generation from \emph{semantic layouts}. In this demo, we progressively build a car by sketching its parts, proposing an interesting way of controlling the generation process.}
\label{fig:semantic-generation}
\vspace{-4mm}
\end{figure}

\paragraph{Semantic mesh generation} Since our framework predicts a 3D semantic layout for each image, we can condition the generator on such a representation. In \autoref{fig:semantic-generation}, we propose a proof-of-concept where we train a conditional model on the \emph{car} class that takes as input a semantic layout in UV space and produces a textured mesh. %
Such a setting can be used to manipulate fine details (e.g.\ the shape of the headlights) or the placement of semantic parts.

\vspace{-1mm}
\section{Conclusion}
\label{sec:conclusion}
\vspace{-1mm}
We proposed a framework for learning generative models of textured 3D meshes. In contrast to prior work, our approach does not require keypoint annotations, enabling its use on real-world datasets.
We demonstrated that our method matches the results of prior works that use ground-truth keypoints, without having to rely on such information. Furthermore, we set new baselines on a subset of categories from ImageNet \cite{deng2009imagenet}, where keypoints are \emph{not} available.
We believe there are still many directions of interest to pursue as future work. In addition to further analyzing disentanglement and exploring more intuitive semantic generation techniques, it would be interesting to experiment with articulated meshes %
and work with more data.%

\paragraph{Acknowledgments} This work was partly supported by the Swiss National Science Foundation (SNF), grant \#176004.

{\small
\bibliographystyle{ieee_fullname}
\bibliography{egbib}
}

\clearpage
\newpage
\appendix

\section{Supplementary material}

\subsection{Implementation details}
\label{sec:appendix-implementation-details}

\paragraph{Dataset preparation} We infer object silhouettes using PointRend \cite{kirillov2020pointrend} with an X101-FPN backbone, using their pretrained model on COCO \cite{lin2014mscoco}. We set the object detection threshold to 0.9 to select only confident objects. As mentioned in \autoref{sec:method}, we discard object instances that are either \emph{(i)} too small (mask area $< 96^2$ pixels), \emph{(ii)} touch the borders of the image (indicator of possible truncation), or \emph{(iii)} collide with other detected objects (indicator of potential occlusion). 
For the object part segmentations, we use the semi-supervised object detector from \cite{hu2018segmenteverything}, which can segment all 3000 classes available in Visual Genome (VG) \cite{krishna2017visualgenome} while being supervised only on mask annotations from COCO. Although this model was not conceived for object part segmentation, we find that it can be used as a cost-effective way of obtaining meaningful part segmentations without collecting extra data or using co-part segmentation models that require class-specific hyperparameter tuning, such as \ \emph{SCOPS} \cite{hung2019scops}. Specifically, since VG presents a long tail of rare classes, as in \cite{pavllo2020stylesemantics} we found it beneficial to first pre-select a small number of representative classes that are widespread across categories (e.g.\ all land vehicles have wheels, all animals have legs). We set the detection threshold of this model to 0.2 and, for each image category, we only keep semantic classes that appear in at least 25\% of the images, which helps eliminate spurious detections. On our data, this leads to a number of semantic classes $K \approx 10$ per image category (33 across all categories). The full list of semantic classes can be seen in \autoref{fig:semantic-templates}. To deal with potentially overlapping part detections (e.g.\ the segmentation mask of the door of a car might overlap with a window), the output semantic maps represent probability distributions over classes, where we weight each semantic class proportionally to the object detection score. Additionally, we add an extra class for ``no class'' (depicted in gray in our figures).

\begin{figure}[ht]
\begin{center}
  \includegraphics[width=\linewidth]{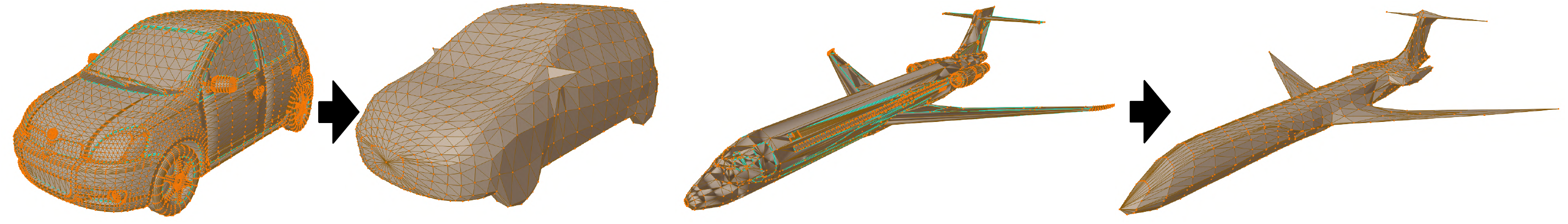}
\end{center}
   \caption{Remeshing of the mesh templates. In this figure we show two demos (one template for \emph{car} and one for \emph{airplane}).}
\label{fig:appendix-remeshing}
\end{figure}

\paragraph{Mesh templates and remeshing} We borrow a selection of mesh templates from \cite{kulkarni2020acsm} as well as meshes freely available on the web. In the experiments where we adopt multiple mesh templates, we only use 2--4 meshes per category. An important preliminary step of our approach, which is performed even before the pose estimation step, consists in \emph{remeshing} these templates to align them to a common topology. This has the goal of reducing their complexity (which translates into a speed-up during optimization), removing potential invisible interiors, and enabling efficient batching by making sure that every mesh has the same number of vertices/faces. Additionally, as mentioned in \autoref{sec:generation-estimation}, remeshing is required for the semi-supervision loss term in the reconstruction model. We frame this task as an optimization problem where we deform a $32\times32$ UV sphere to match the mesh template. More specifically, we render each template from 64 random viewpoints at $256 \times 256$ resolution, and minimize the MSE loss between the rendered deformed sphere and the target template in pixel space ($\mathcal{L}_\text{MSE}$). Moreover, we regularize the mesh by adding \emph{(i)} a smoothness loss $\mathcal{L}_\text{flat}$, which encourages neighboring faces to have similar normals, \emph{(ii)} a Laplacian smoothing loss $\mathcal{L}_\text{lap}$ with quad connectivity (i.e.\ using the topology of the UV map as opposed to that of the triangle mesh), and \emph{(iii)} an edge length loss $\mathcal{L}_\text{len}$ with quad connectivity, which encourages edges to have similar lengths. $\mathcal{L}_\text{flat}$ and $\mathcal{L}_\text{len}$ are defined as follows:
\setlength{\belowdisplayskip}{6pt}
\setlength{\abovedisplayskip}{6pt}
\begin{equation}
\resizebox{0.50\linewidth}{!}{$\displaystyle\mathcal{L}_\text{flat} = \frac{1}{|E|} \sum_{i, j \in E} (1 - \cos\theta_{ij})^2$}
\end{equation}
\begin{equation}
    \resizebox{0.9\linewidth}{!}{$\displaystyle\mathcal{L}_\text{len} = \frac{1}{|UV|} \sum_{i \in U}  \sum_{j \in V} \frac{\lVert \mathbf{v}_{i+1, j} - \mathbf{v}_{i, j} \rVert_1 + \lVert \mathbf{v}_{i, j+1} - \mathbf{v}_{i, j} \rVert_1}{6}$}
\end{equation}
where $E$ is the set of edges, $\cos\theta_{ij}$ is the cosine similarity between the normals of faces $i$ and $j$, and $\mathbf{v}_{i,j}$ represents the 3D vertex at the coordinates $i, j$ of the UV map.\\
Finally, we weight each term as follows:
\begin{equation}
    \mathcal{L} = \mathcal{L}_\text{MSE} + 0.00001\,\mathcal{L}_\text{flat} + 0.003\,\mathcal{L}_\text{lap} + 0.01\,\mathcal{L}_\text{len}
\end{equation}
Additionally, in the experiments with multiple mesh templates, we add a pairwise similarity loss $\mathcal{L}_\text{align}$ which penalizes large variations of the vertex positions between different mesh templates (only within the same category):
\begin{align}
    \mathcal{L}_\text{align} = \frac{1}{N_t^2} \sum_{i = 1}^{N_t} \sum_{j = 1}^{N_t} \lVert \mathbf{V_i} - \mathbf{V_j} \rVert_2
\end{align}
where $\mathbf{V_i}$ is a matrix that contains the vertex positions of the $i$-th mesh template (of shape $3 \times N_v$), and $N_t$ is the number of mesh templates. This loss term is added to the total loss with weight $0.001$. Note that we use a non-squared L2 penalty for this term, which encourages a sparse set of vertices to change between mesh templates.\\
We optimize the final loss using SGD with momentum (initial learning rate $\alpha = 0.0001$ and momentum $\beta = 0.9$). We linearly increase $\alpha$ to $0.0005$ over the course of 500 iterations (warm-up) and then exponentially decay $\alpha$ with rate $0.9999$. We stop when the learning rate falls below $0.0001$. Additionally, we normalize the gradient before each update. \autoref{fig:appendix-remeshing} shows two qualitative examples of remeshing.

\paragraph{Pose estimation} For the silhouette optimization step, we initialize $N_c = 40$ camera hypotheses per image by uniformly quantizing azimuth and elevation (8 quantization levels along azimuth and 5 levels along elevation). We optimize each camera hypothesis using Adam \cite{kingma2014adam} with full-matrix preconditioning, where we set $\beta_1 = 0.9$ and $\beta_2 = 0.95$. %
The implementation of our variant of Adam as well its theoretical justification are described in the next paragraph. We optimize each hypothesis for 100 iterations, with an initial learning rate $\alpha = 0.1$ which is decayed to 0.01 after the 80th iteration. After each iteration, we reproject quaternions onto the unit ball. As a performance optimization, silhouettes are initially rendered at 128$\times$128 resolution, which is increased to 192$\times$192 after the 30th iteration and 256$\times$256 after the 60th iteration. Finally, in the settings where we prune camera hypotheses, we discard the worst 50\% hypotheses as measured by the intersection-over-union (IoU) between projected and target silhouettes. This is performed twice: after the 30th and 60th iteration.

\renewcommand{\algorithmiccomment}[1]{\bgroup\hfill $\triangleright$ ~#1\egroup}
\newcommand{\vc}[1]{\textit{\textbf{#1}}}
\newcommand{\cb}[1]{\colorbox{ForestGreen}{#1}}
\begin{algorithm}[ht]
\small
\caption{Adam with full-matrix preconditioning.\\Changes w.r.t. the original algorithm are \cb{highlighted}.}
\label{alg:appendix-adam-full}
\begin{algorithmic}[1]
\STATE{\textbf{require} $\alpha$ (step size), $\beta_1, \beta_2, \epsilon$}
\STATE{\textbf{initialize} time step $t \leftarrow 0$}
\STATE{\textbf{initialize} parameters $\boldsymbol{\theta}_{0}$ ($d$-dimensional col.\ vector)}
\STATE{\textbf{initialize} first moment $\vc{m}_{0} \leftarrow \vc{0}$ ($d$-dimensional col.\ vector)}
\STATE{\textbf{initialize} second moment \cb{$\vc{V}_{0}$}$\leftarrow \vc{0}$ ($d \times d$ \cb{matrix})}
\REPEAT
	\STATE{$t \leftarrow t + 1$}
	\STATE{$\vc{g}_t \leftarrow \nabla_{\boldsymbol{\theta}} f_t(\boldsymbol{\theta}_{t-1})$} \COMMENT{gradient}
	\STATE{$\vc{m}_t \leftarrow \beta_1 \vc{m}_{t-1} + (1 - \beta_1) \vc{g}_t $} \COMMENT{first moment}
	\STATE{$\vc{V}_t \leftarrow \beta_2 \vc{V}_{t-1} + (1 - \beta_2)$ \cb{$\vc{g}_t\vc{g}^T_t$}} \COMMENT{second moment}
	\STATE{$\hat{\vc{m}}_t \leftarrow \vc{m}_t/(1 - \beta_1^t) $} \COMMENT{bias correction}
	\STATE{$\hat{\vc{{V}}}_t \leftarrow \vc{V}_t/(1 - \beta_2^t) $} \COMMENT{bias correction}
	\STATE{$\boldsymbol{\theta}_t \leftarrow \boldsymbol{\theta}_{t-1} - \alpha $ \cb{$(\hat{\vc{V}}_t + \epsilon \mathbf{I}_d)^{-\frac{1}{2}}$}$\hat{\vc{m}}_t$} \COMMENT{update}
\UNTIL{\textit{stopping criterion} }
\RETURN{$\boldsymbol{\theta}_t$}
\end{algorithmic}
\end{algorithm}

\paragraph{Full-matrix preconditioning} Adam \cite{kingma2014adam} is an established optimizer for training neural networks. Its use of diagonal preconditioning is an effective  trick to avoid storing an $\mathcal{O}(d^2)$ matrix for the second moments (where $d$ is the number of learnable parameters), for which a matrix square root and inverse need to be subsequently computed (an extra $\mathcal{O}(d^3)$ cost for each of the two operations). However, since our goal is to optimize camera parameters, we observe that:
\begin{enumerate}[leftmargin=*, itemsep=-2pt]
    \item Optimizers with diagonal preconditioning are not rotation invariant, i.e.\ they have some preferential directions that might bias the pose estimation result.
    \item Since each camera hypothesis comprises only 8 parameters, inverting an $8 \times 8$ matrix has a negligible cost.
\end{enumerate}
Using a rotation invariant optimizer such as SGD (with or without momentum) is a more principled choice as it addresses the first observation. However, based on our second observation, we take the best of both worlds and modify Adam to implement full-matrix preconditioning. This only requires a trivial modification to the original implementation, which we show in \autoref{alg:appendix-adam-full} (changes w.r.t. the original algorithm are highlighted in green).

\begin{figure}[ht]
\begin{center}
  \includegraphics[width=\linewidth]{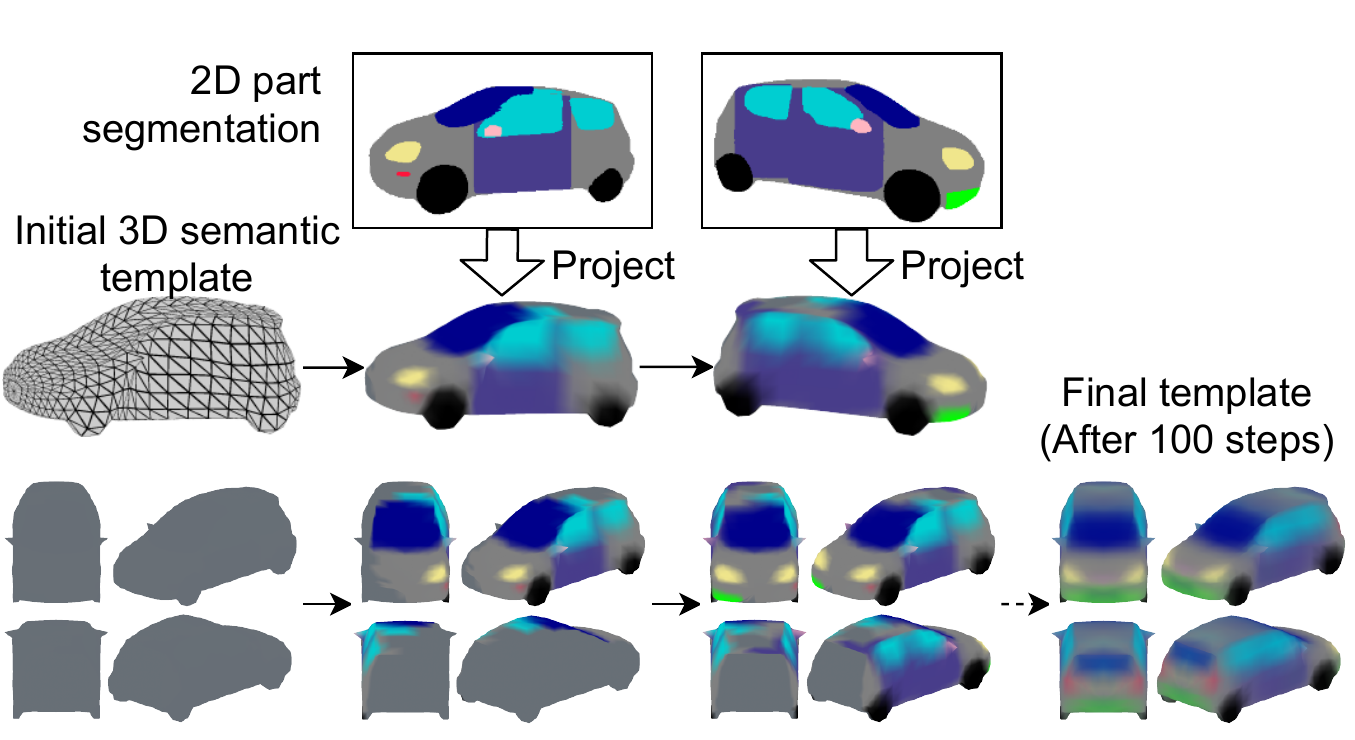}
\end{center}
   \caption{Semantic template inference, starting from an untextured 3D mesh template (left-to-right progression). In this figure we show a demo with two sample images, and the final result using the top 100 images as measured by the IoU.}
\label{fig:appendix-semantic-inference}
\end{figure}

\paragraph{Semantic template inference} As mentioned in \autoref{sec:pose-estimation}, the goal of this step is to infer a 3D semantic template for each mesh template, given an initial (untextured) mesh template, the output of the silhouette optimization step, and a collection of 2D semantic maps. Recapitulating from \autoref{sec:pose-estimation}, we solve the following optimization problem:
\begin{align}
    \mathcal{L}_i &= \left\lVert \mathcal{R}(\mathbf{V}_\text{tpl}, \mathbf{F}_\text{tpl}, \mathbf{C}_\text{tpl};\; \mathbf{q_i}, \mathbf{t_i}, s_i, {z_0}_i) - \mathbf{C_i} \right\rVert^2 \\
    \mathbf{C}_\text{tpl}^* &= \min_{\mathbf{C}_\text{tpl}} \frac{1}{N_\text{top}} \sum_i \mathcal{L}_i
\end{align}
Conceptually, our goal is to learn a shared semantic template (parameterized using vertex colors) that averages all 2D semantic maps in vertex space. We propose the following closed-form solution which uses the gradients from the differentiable renderer and requires only a single pass through the dataset:
\begin{align}
    \mathbf{A} &= \sum_i \nabla_{\mathbf{C}_\text{tpl}} (\mathcal{L}_i)\\
    (\mathbf{C}_\text{tpl}^*)_{k} &= \frac{\epsilon +\mathbf{a_{k}} }{K\epsilon + \sum_j \mathbf{a_{j}}}
\end{align}
where $\mathbf{A}$ is an accumulator matrix that has the same shape as the $\mathbf{C}_\text{tpl}$ (the vertex colors), and $\epsilon$ is a small \emph{additive smoothing} constant that leads to a uniform distribution on vertices that are never rendered (and thus have no gradient). This operation can be regarded as projecting the 2D object-part semantics onto the mesh vertices and computing a color histogram on each vertex. We show a sample illustration in \autoref{fig:appendix-semantic-inference}.

In section \autoref{sec:pose-estimation} we explained that we compute the semantic template using the top $N_\text{top} = 100$ images as measured by the IoU, among those that passed the ambiguity detection test ($v_\text{agr} < 0.3$). To further improve the quality of the inferred semantic templates, we found it beneficial to add an additional filter where we only select poses whose cosine distance is within 0.5 (i.e.\ 45 degrees) of the left/right side. Objects observed from the left/right side are intrinsically unambiguous, since there is no complementary pose that results in the same silhouette. Therefore, we favor views that are close to the left/right as opposed to the front/back or top/bottom, which are the most ambiguous views. Note that this filter is only used for the semantic template inference step. %

\paragraph{Generative model} We train the single-category reconstruction networks (setting \textbf{A}) for 130k iterations, with a batch size of 32, and on a single GPU. The multi-category model (setting \textbf{B}) is trained for 1000 epochs, with a total batch size of 128 across 4 GPUs, using synchronized batch normalization. In both settings, we use Adam \cite{kingma2014adam} (the original one, not our variant with full-matrix preconditioning) with an initial learning rate of $0.0001$ which is halved at 1/4, 1/2, 3/4 of the training schedule. For the GAN, we use the same hyperparameters as \cite{pavllo2020convmesh}, except in the multi-category model (setting \textbf{B}), which is trained with a batch size of 64 instead of the default 32. Furthermore, in setting \textbf{B}, and for both models (reconstruction and GAN), we equalize classes during mini-batch sampling. This is motivated by the large variability in the amount of training images, as explained in \autoref{sec:datasets}, and as can also be seen in \autoref{tab:appendix-dataset-stats}.
Finally, as in \cite{pavllo2020convmesh, kanazawa2018cmr, goel2020ucmr, li2020umr}, we force generated meshes to be left/right symmetric.

\paragraph{Semantic mesh generation} In the setting where we generate a 3D mesh from a semantic layout in UV space, we modify the generator architecture of \cite{pavllo2020convmesh}. Specifically, we replace the input linear layer (the one that projects the latent code $\mathbf{z}$ onto the first $8 \times 8$ convolutional feature map) with four convolutional layers. These progressively downsample the semantic layout from $128 \times 128$ down to $8 \times 8$ (i.e.\ each layer has stride 2). The first layer takes as input a \emph{one-hot} semantic map (with $K$ semantic channels) and yields 64 output channels (128, 256, 512 in the following layers). In these 4 layers, we use Leaky ReLU activations (slope 0.2), spectral normalization, but no batch normalization. We leave the rest of the network unchanged. In this model, we also found it necessary to fine-tune the batch normalization statistics prior to evaluation, which we do by running a forward pass over the entire dataset on the \emph{running average} model. As for the discriminator, we simply resize the semantic map as required and concatenate it to the input.

\subsection{Additional results}
\label{sec:appendix-additional-results}

\begin{figure}[ht]
\begin{center}
  \includegraphics[width=0.49\linewidth]{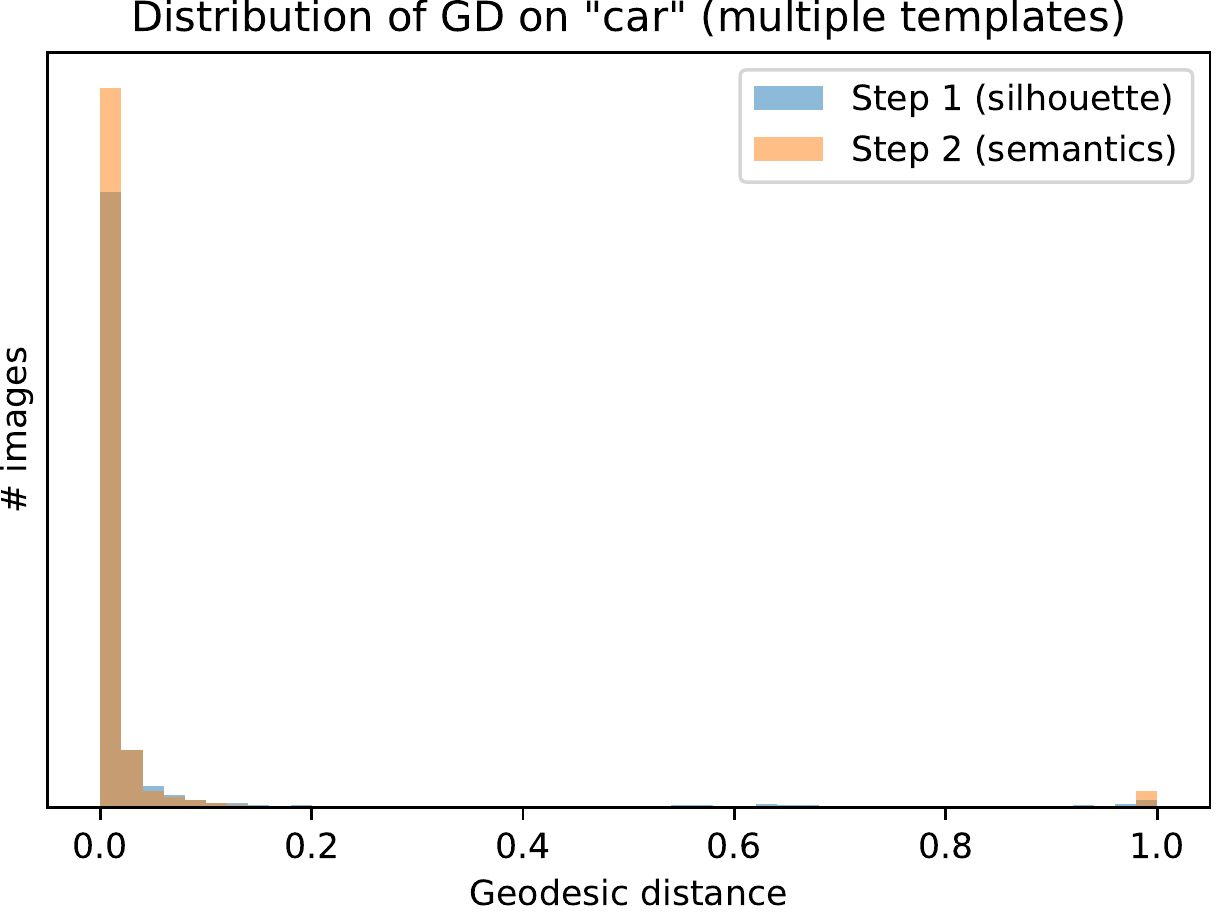}
  \includegraphics[width=0.49\linewidth]{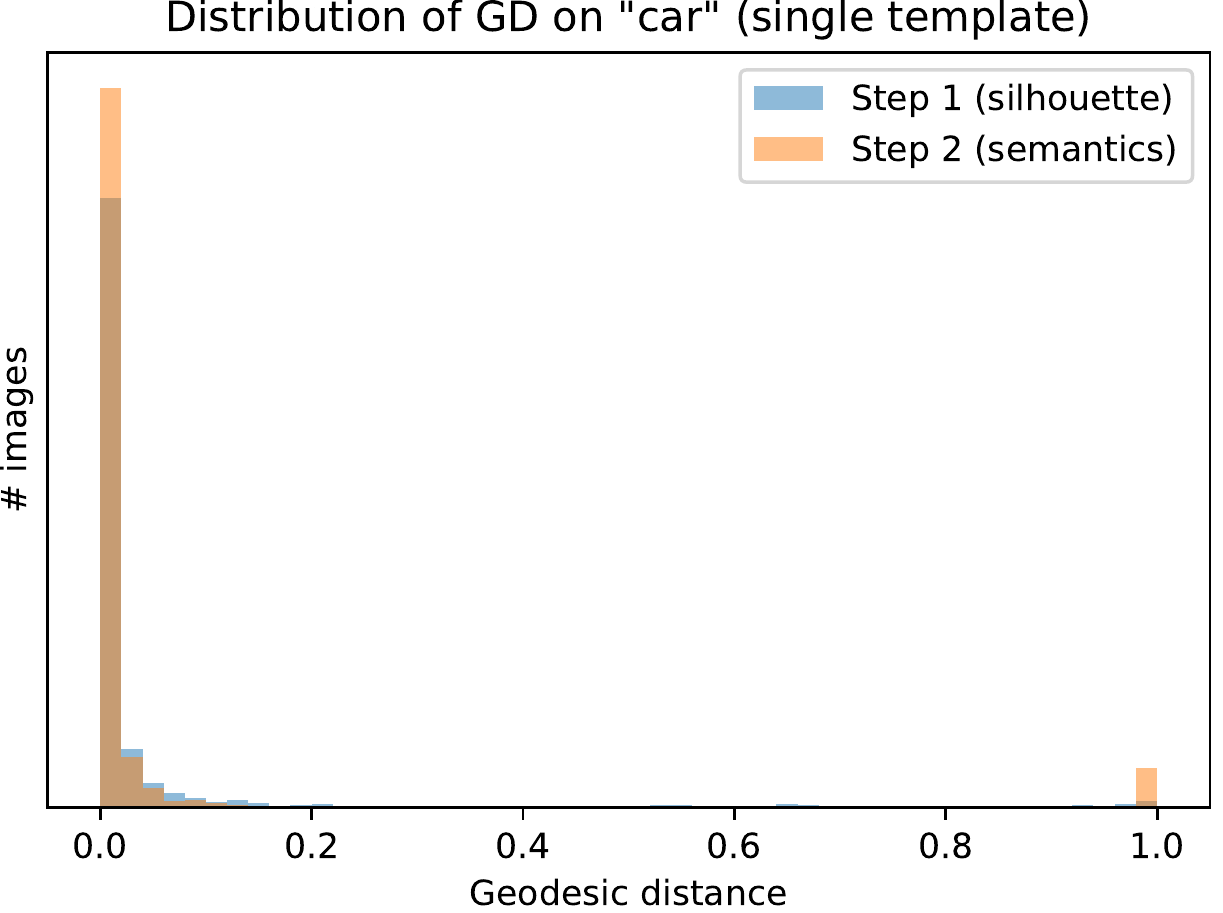}\\
  \vspace{2mm}
  \includegraphics[width=0.49\linewidth]{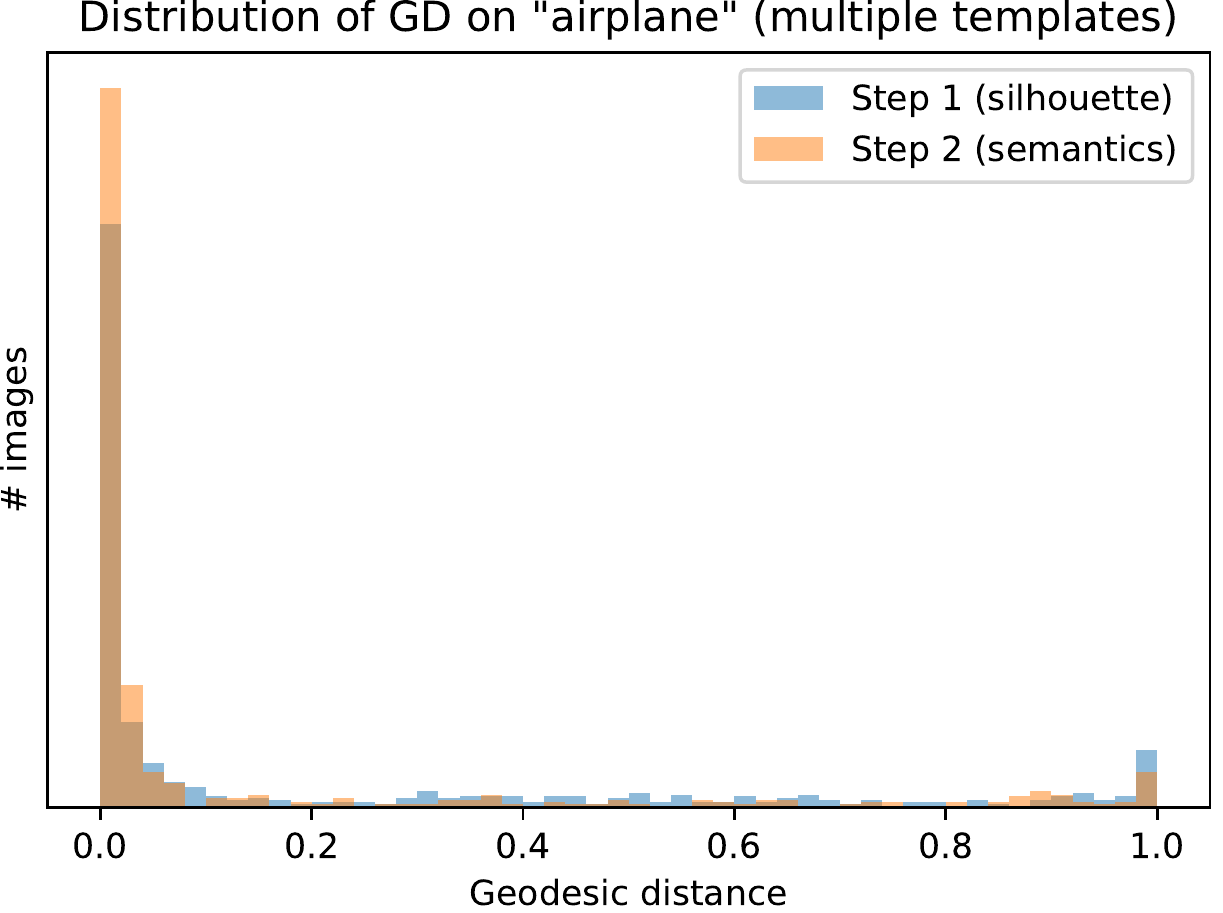}
  \includegraphics[width=0.49\linewidth]{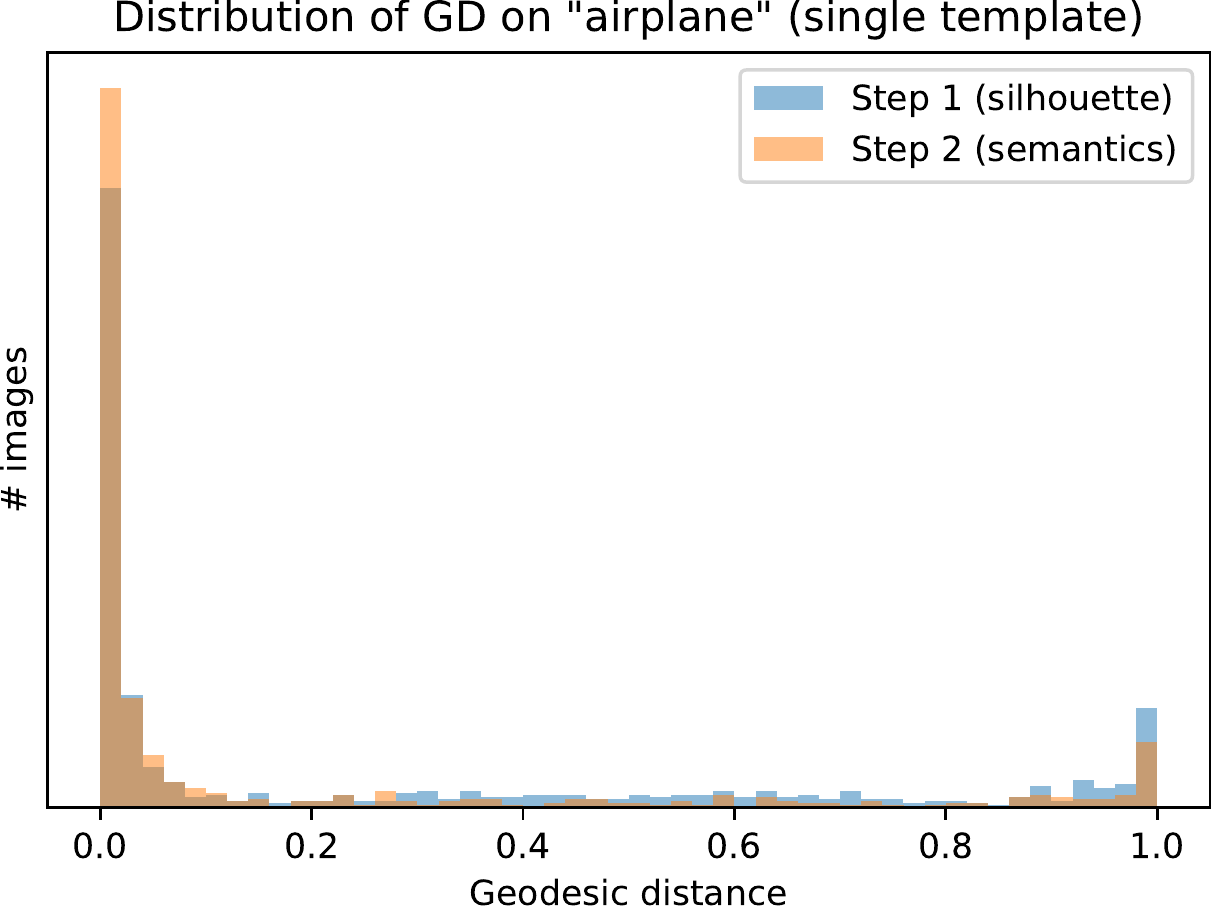}\\
  \vspace{2mm}
  \includegraphics[width=0.49\linewidth]{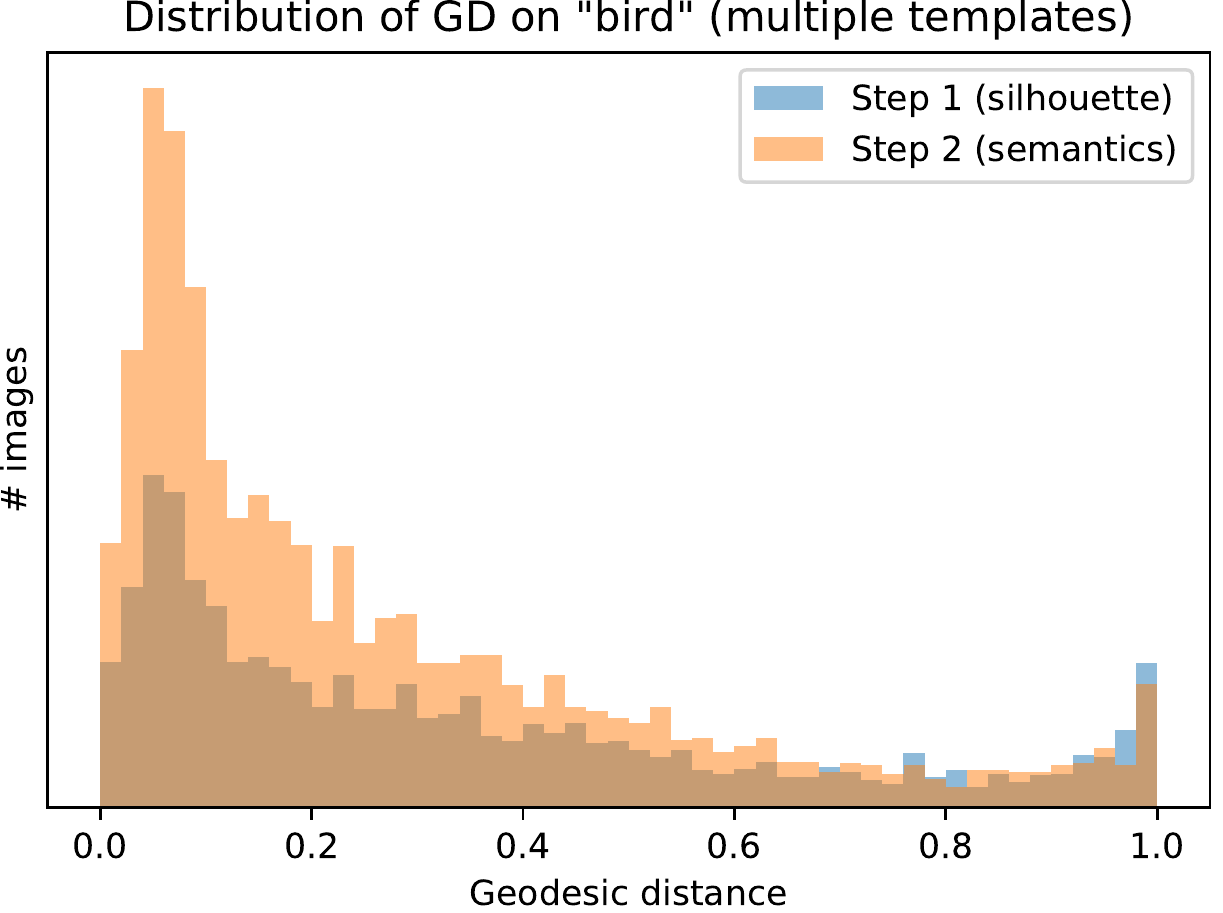}
  \includegraphics[width=0.49\linewidth]{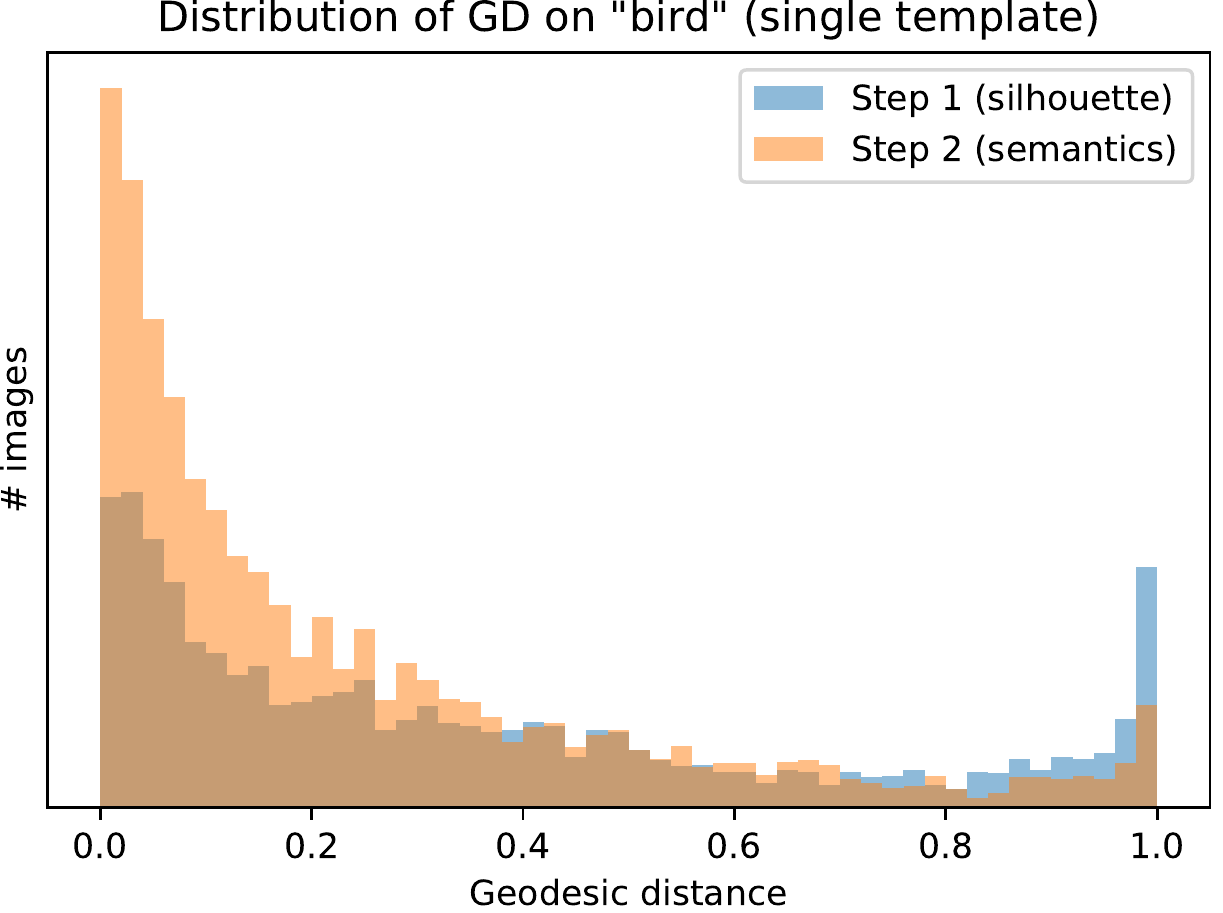}
\end{center}
   \caption{Distribution of pose estimation errors on \emph{car}, \emph{airplane}, and \emph{bird}. We compare settings where we use multiple mesh templates (left) and a single template (right).}
\label{fig:appendix-pose-estimation}
\end{figure}

\paragraph{Pose estimation} In \autoref{fig:appendix-pose-estimation}, we provide more insight into the \emph{geodesic distance} metric, which measures the cosine distance between the rotations predicted by our approach (\autoref{sec:pose-estimation}) and SfM rotations. In particular, as opposed to the results presented in \autoref{tab:pose-estimation-results} (which shows only the average), here we show the full distribution of errors. A distance of $0$ means that the two rotations match exactly, whereas a distance of $1$ (maximum value) means that the rotations are rotated by 180 degrees from one another. On the analyzed classes  (\emph{car}, \emph{airplane}, and \emph{bird}, for which we have SfM poses), we can generally observe a bimodal distribution: a majority of images where pose estimation is correct, i.e.\ the GD is close to zero, and a small cluster of images where the GD is close to one. This is often the case for ambiguities: for instance, in cars we sometimes observe a front/back confusion. As expected, exploiting semantics (step 2) mitigates this issue and increases the amount of available images (this is particularly visible on \emph{bird}). We also note that, for rigid objects such as \emph{car} and \emph{airplane}, the distribution is more peaky, whereas for \emph{bird} the tail of errors is longer, most likely because pose estimation is more ill-defined for articulated objects.

\begin{figure}[t]
\begin{center}
  \includegraphics[width=\linewidth]{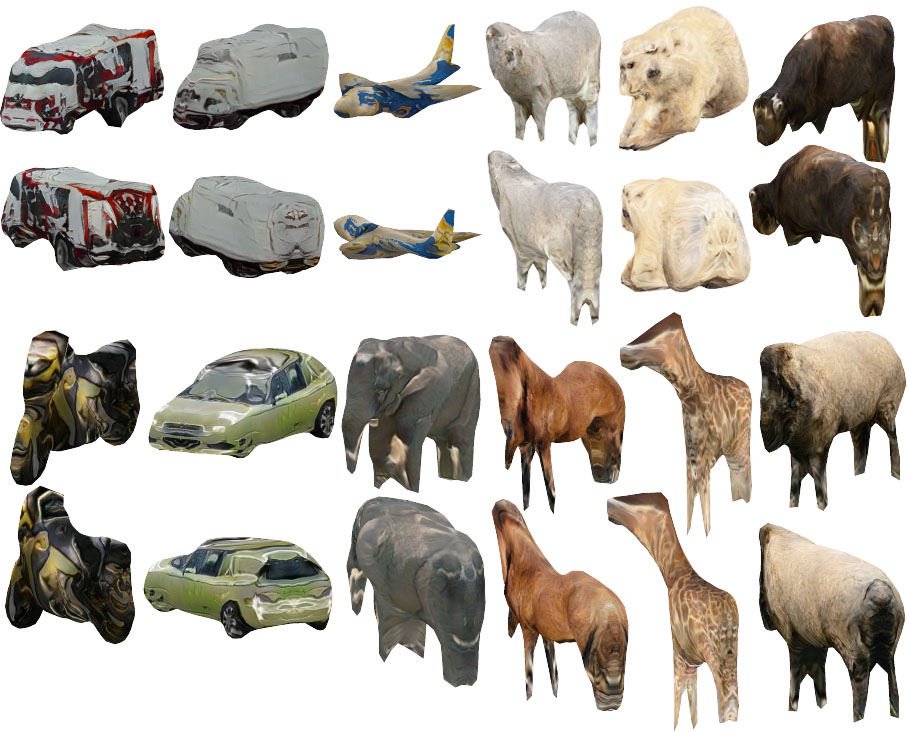}
\end{center}
   \caption{Failure cases for a variety of categories.}
\label{fig:appendix-failure-cases}
\end{figure}

\begin{figure*}[p]
\begin{center}
  \includegraphics[width=\textwidth]{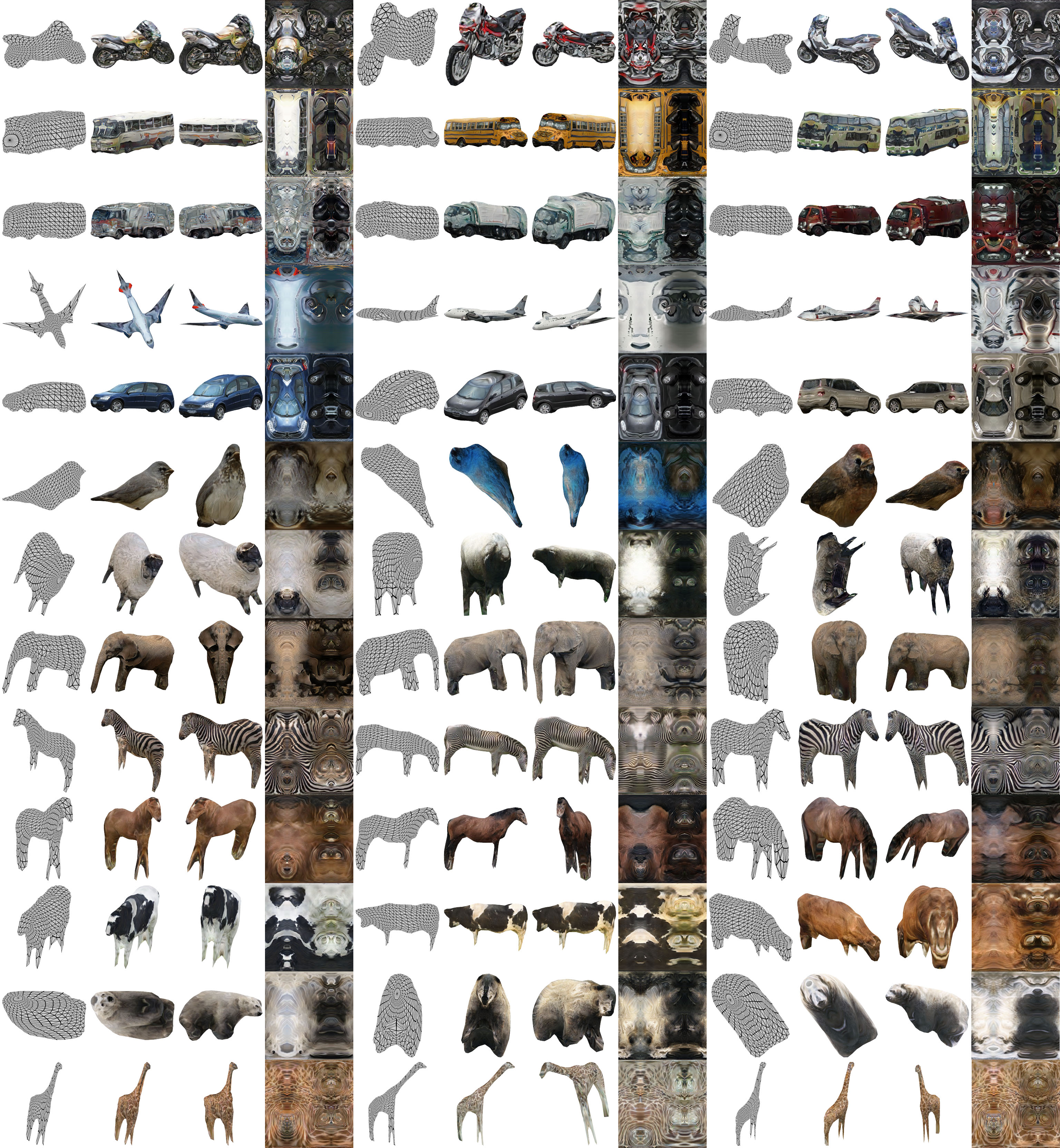}
\end{center}
   \caption{Additional qualitative results. We show three examples per category. Each example is rendered from two random views, and the corresponding texture/wireframe mesh is also shown.}
\label{fig:appendix-qualitative}
\end{figure*}

\paragraph{Qualitative results} We show extra qualitative results in \autoref{fig:appendix-qualitative}. In particular, we render each generated mesh from two random viewpoints and showcase the associated texture and wireframe mesh. Additionally, in \autoref{fig:appendix-failure-cases} we show the most common failure cases across categories. We can identify some general patterns: for instance, in vehicles we sometimes observe incoherent textures (this is particularly visible in \emph{truck} due to the small size of this dataset). On animals, as mentioned, we observe occasional failures to model facial details, merged/distorted legs, and more rarely, mesh distortions. To some extent,  these issues can be mitigated by sampling from the generator using a lower truncation threshold (we use $\sigma = 1.0$ in our experiments), at the expense of sample diversity.

\begin{figure*}[t]
\begin{center}
  \includegraphics[width=\textwidth]{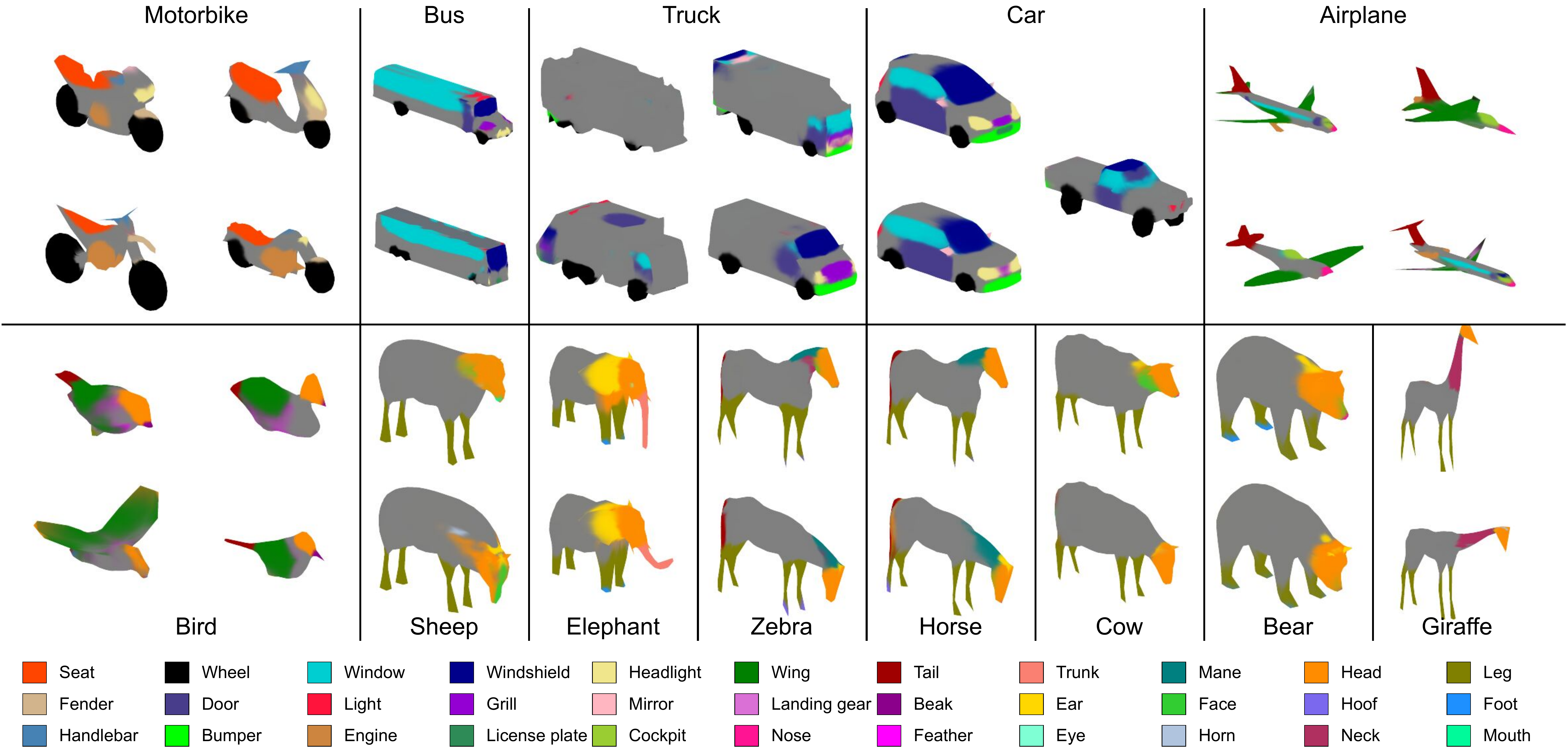}
\end{center}
   \caption{Visualization of \emph{all} the learned 3D semantic templates (2--4 per category). While most results are as expected, the figure highlights some failure cases, e.g.\ in \emph{truck} some templates have very few images assigned to them, which leads to incoherent semantics.}
\label{fig:appendix-semantic-templates-full}
\end{figure*}

\paragraph{Semantic templates} \autoref{fig:appendix-semantic-templates-full} shows the full set of learned semantic templates for every category. Most results are coherent, although we observe a small number of failure cases, e.g.\ in \emph{truck} one or two templates are mostly empty and are thus ineffective for properly resolving ambiguities. This generally happens when the templates have too few images assigned to them and explains why the multi-template setting does not consistently outperform the single-template setting.

\paragraph{Demo video} The supplementary material includes a video where we show additional qualitative results. First, we showcase samples generated by our models in setting \textbf{A} and explore the latent space of the generator. Second, we analyze the latent space of the model trained to generate multiple classes (setting \textbf{B}), and discover interpretable directions in the latent space, which can be used to control shared aspects between classes (e.g.\ lighting, shadows). We also interpolate between different classes while keeping the latent space fixed, and highlight that style is preserved during interpolation. Finally, we showcase a setting where we generate a mesh from a hand-drawn semantic layout in UV space, similar to \autoref{fig:semantic-generation}.

\subsection{Dataset information}
\label{sec:appendix-dataset-information}

For our experiments on ImageNet, we adopt the \emph{synsets} specified in \autoref{tab:appendix-dataset-stats}. Since some of our required synsets are not available in the more popular ImageNet1k, we draw all of our data from the larger ImageNet22k set. 

\begin{table}[t]
\begin{center}
\resizebox{\linewidth}{!}{
\begin{tabular}{l|l|l|l}
Class      & Synsets                                                                                                                                                                                   & Raw images & Valid instances \\ \hline\hline
Motorbike & \begin{tabular}[c]{@{}l@{}}n03790512,      n03791053,      n04466871\end{tabular}                                                                                                       & 4037          & 1351               \\ \hline
Bus        & \begin{tabular}[c]{@{}l@{}}n04146614,      n02924116\end{tabular}                                                                                                                        & 2641          & 1190               \\ \hline
Truck      & \begin{tabular}[c]{@{}l@{}}n03345487,      n03417042,      n03796401\end{tabular}                                                                                                       & 3187          & 1245               \\ \hline
Car        & \begin{tabular}[c]{@{}l@{}}n02814533,      n02958343,      n03498781,\\      n03770085,      n03770679,      n03930630,\\      n04037443,      n04166281,      n04285965\end{tabular} & 12819         & 4992               \\ \hline
Airplane   & \begin{tabular}[c]{@{}l@{}}n02690373,      n02691156,      n03335030,\\      n04012084\end{tabular}                                                                                      & 5208          & 2540               \\ \hline
Sheep      & \begin{tabular}[c]{@{}l@{}}n10588074,      n02411705,      n02413050,\\      n02412210\end{tabular}                                                                                      & 4682          & 864                \\ \hline
Elephant   & \begin{tabular}[c]{@{}l@{}}n02504013,      n02504458\end{tabular}                                                                                                                        & 3927          & 1434               \\ \hline
Zebra      & \begin{tabular}[c]{@{}l@{}}n02391049,      n02391234,      n02391373,\\      n02391508\end{tabular}                                                                                      & 5536          & 1753               \\ \hline
Horse      & \begin{tabular}[c]{@{}l@{}}n02381460,      n02374451\end{tabular}                                                                                                                        & 2589          & 664                \\ \hline
Cow        & \begin{tabular}[c]{@{}l@{}}n01887787,      n02402425\end{tabular}                                                                                                                        & 2949          & 861                \\ \hline
Bear       & \begin{tabular}[c]{@{}l@{}}n02132136,      n02133161,      n02131653,\\      n02134084\end{tabular}                                                                                      & 6745          & 2688               \\ \hline
Giraffe    & n02439033                                                                                                                                                                                 & 1256          & 349               
\end{tabular}
}
\end{center}
\caption{Synsets and summary statistics for our ImageNet data. For each category, we report the number of raw images in the dataset, and the number of extracted object instances that have passed our quality checks (size, truncation, occlusion).}
\label{tab:appendix-dataset-stats}
\end{table}

\subsection{Negative results}
\label{sec:appendix-negative-results}

To guide potential future work in this area, we provide a list of ideas that we explored but did not work out.

\paragraph{Silhouette optimization} For the silhouette optimization step with multiple templates, before reaching our current formulation, we explored a range of alternatives. In particular, we tried to smoothly interpolate between multiple meshes by optimizing a set of interpolation weights along with the camera parameters. This yielded inconsistent results across categories, which convinced us to work with a ``discrete'' approach as opposed to a smooth one. We then tried a reinforcement learning approach inspired by multi-armed bandits: we initialized each camera hypothesis with a random mesh template, and used a UCB (upper confidence bound) selection algorithm to select the optimal mesh template during optimization. This led to slightly worse results than interpolation. Finally, we reached our current formulation, where we simply replicate each camera hypothesis and optimize the different mesh templates separately. We adopt pruning to make up for the increase in computation time.

\paragraph{Re-optimizing poses multiple times} In our current formulation, after the semantic template inference step, we use the semantic templates to resolve ambiguities, but there is no further optimization involved. Naturally, we explored the idea of repeating the silhouette optimization step using semantic information. However, we were unable to get this step to work reliably, even after attempting with multiple renderers (we tried both with DIB-R \cite{chen2019dibr} and SoftRas \cite{liu2019softras}). We generally observed that the color gradients are too uninformative for optimizing camera poses, even after trying to balance the different components of the gradient (silhouette and color). We believe this is a fundamental issue related to the non-convexity of the loss landspace, which future work needs to address. We also tried to smooth out the rendered images prior to computing the MSE loss, without success.

\paragraph{Remeshing} Since target 3D vertices are known in this step, we initially tried to use a 3D chamfer loss to match the mesh template. This, however, led to artifacts and merged legs in animals, and was too sensitive to initialization. We found it more reliable to use a differentiable render with silhouette-based optimization.

\end{document}